\documentclass[sigconf,nonacm]{acmart}

\usepackage{subcaption}
\usepackage{algorithm}
\usepackage{algorithmic}

\usepackage[labelformat=simple]{subcaption}

\newcommand{\mps}[1]{\mathbb{N}^{#1}}
\newcommand{\setplus}{\uplus}

\begin{document}

\title{REACT: Revealing Evolutionary Action Consequence Trajectories for Interpretable Reinforcement Learning} 

\author{Philipp Altmann}
\orcid{0000-0003-1134-176X}
\affiliation{\institution{LMU Munich}\city{Munich}\country{Germany}}
\email{philipp.altmann@ifi.lmu.de}

\author{C\'{e}line Davignon}
\affiliation{\institution{LMU Munich}\city{Munich}\country{Germany}}
\orcid{0009-0005-8935-6293}

\author{Maximilian Zorn}
\affiliation{\institution{LMU Munich}\city{Munich}\country{Germany}}
\orcid{0009-0006-2750-7495}

\author{Fabian Ritz}
\affiliation{\institution{LMU Munich}\city{Munich}\country{Germany}}
\orcid{0000-0001-7707-1358}

\author{Claudia Linnhoff-Popien}
\affiliation{\institution{LMU Munich}\city{Munich}\country{Germany}}
\orcid{0000-0001-6284-9286}

\author{Thomas Gabor}
\affiliation{\institution{LMU Munich}\city{Munich}\country{Germany}}
\orcid{0000-0003-2048-8667}

\renewcommand{\shortauthors}{Altmann et al.}

\begin{abstract}
To enhance the interpretability of Reinforcement Learning (RL), we propose \textit{Revealing Evolutionary Action Consequence Trajectories} (REACT). 
In contrast to the prevalent practice of validating RL models based on their optimal behavior learned during training, we posit that considering a range of edge-case trajectories provides a more comprehensive understanding of their inherent behavior. 
To induce such scenarios, we introduce a disturbance to the initial state, optimizing it through an evolutionary algorithm to generate a diverse population of demonstrations. 
To evaluate the fitness of trajectories, REACT incorporates a joint fitness function that encourages both local and global diversity in the encountered states and chosen actions. 
Through assessments with policies trained for varying durations in discrete and continuous environments, we demonstrate the descriptive power of REACT. 
Our results highlight its effectiveness in revealing nuanced aspects of RL models' behavior beyond optimal performance, thereby contributing to improved interpretability. 
\end{abstract}



\maketitle

\section{Introduction}

With the increasing use of large, parameterized function approximation models, there is a growing demand for interpretation methods that bridge the gap between human understanding and computational intelligence. 
This need is particularly pronounced in the context of complex dynamic approaches like Reinforcement Learning (RL), where existing validation methods primarily focus on optimal behavior. 

In high-dimensional (continuous) state and action spaces, RL policies are usually realized with parameterized neural networks.
This results in a number of challenges:
First, contrary to static supervised learning tasks like classification, RL policies are inherently hard to visualize, especially given the intended application to varying circumstances.
Second, demonstrating the desired behavior in a laboratory training setup does not serve as sufficient validation to enable the interpretability of the inherent behavior. 
Third, \textit{comparative evaluation} is often considered to play a central role in the comprehension, explanation, and interpretation of varying phenomena by providing additional context information and, thus, control~\cite{vartiainen2002principles}.

To tackle these challenges, we propose to evaluate a set of diverse edge-case demonstrations which we obtain by precisely disturbing the initial state.
As we expect a human (expert) to perform the evaluation, we have to consider fatigue~\cite{behrens2023fatigue}, which constrains the number of demonstrations.
To generate a small, yet informative set of demonstrations, we employ evolutionary optimization, which can be adapted to yield diverse solution candidates in complex solution landscapes across various (local) optima.
To harness these prospects, we propose a framework to indirectly optimize a population of demonstration behavior generated by a given (trained) policy by altering (disturbing) the initial state. 
Overall, we provide the following contributions:

\begin{itemize}
\item We formalize a joint fitness metric to assess demonstration trajectories w.r.t. their local (inherent) and global (comparative) state diversity and action certainty. 
\item We propose an architecture for \textit{Revealing Evolutionary Action Consequence Trajectories} (REACT), integrating the previously defined fitness to optimize a pool of diverse demonstrations to serve as a basis for interpreting the underlying policy.
\item We evaluate the framework in flat and holey gridworlds, as well as a continuous robotic control task, comparing policies of different training stages. 
\end{itemize}

\section{Background}

\subsection{Reinforcement Learning}

We focus on problems formalized as finite-horizon \textit{Markov decision processes} (MDPs) $M=\langle\mathcal{S},\mathcal{A},\mathcal{P},\mathcal{R},\mu,\gamma\rangle$, with a set $\mathcal{S}$ of states $s$, a set $\mathcal{A}$ of actions $a$, a transition probability $\mathcal{P}(s'\mid s, a)$ of reaching $s'$ when executing $a$ in $s$, a scalar reward $r_t = \mathcal{R}(s,a,s')\in\mathbb{R}$ at step $t$, the initial state distribution $s_0\sim\mu$, and the discount factor $\gamma\in[0,1)$ for calculating the discounted return $G_t=\sum_{k=0}^{\infty}\gamma^k r_{t+k}$ \cite{puterman1990markov}.
More specifically, we consider learning in a constrained setting with a single deterministic initial state $s^\star_0$ and evaluating with initial states drawn from $\mu$.
Furthermore, we consider the objective of \textit{reinforcement learning} (RL) to find an optimal policy $\pi^*$ with action selection probability $\pi(a\mid s)$ that maximizes the expected discounted return \cite{suttonbarto01}. 
The \textit{value function} $Q_\pi(s,a)=\mathbb{E}_{\pi}[G_t \mid s,a]$, and the \textit{state-value function} $V_\pi(s)=\mathbb{E}_{\pi,\mathcal{P}}[G_t \mid s]$ evaluate the current policy. 
\textit{Value-based} approaches like Q-learning derive a policy from the constantly updated $Q$ through multi-armed bandit selection \cite{suttonbarto01}. 
In low-dimensional discrete state and action spaces, a tabular representation according to update rules can be sufficient to learn the intended behavior. 
To enable learning in complex high-dimensional or continuous scenarios, using more scaleable models like function approximations is necessary.
We, therefore, consider $\pi$, $V_\pi$, and $Q_\pi$ to be realized by deep neural networks, which are parameterized by a weight vector $\theta$.
However, for simplicity, we omit $\theta$ in the following.
\textit{Policy-based} methods directly approximate the optimal policy from trajectories $\tau$ of experience tuples $\langle s,a,r,s'\rangle$ generated by $\pi$.
Actor-critic methods enable learning from incomplete episodes by approximating future returns using $V$ (the critic) to update $\pi$ (the actor). 
\textit{Proximal policy optimization} (PPO) extends this concept, optimizing a surrogate loss that restricts policy updates to improve the robustness \cite{schulman2017proximal}. 
\textit{Soft actor-critic} (SAC) bridges the gap between value-based Q-learning and policy-based actor-critic methods \cite{SAC}. 
Both algorithms have shown versatile applicability to various scenarios.
Thus, we use both approaches to train the policies on which we base our empirical studies.

\subsection{Explainability}
While enabling learning in complex scenarios, using deep neural networks to approximate the optimal policy comes at the cost of introducing a black-box model into the process.
Therefore, even when finding a parameterization that resembles an optimal policy, its decision cannot be anticipated and reasons for action choices cannot be (easily) inferred.
Yet, RL has been proposed to provide compelling solutions to various real-world decision-making problems such as autonomous driving or robotic control \cite{wurman2022outracing,fu2020d4rl}.
Such problems require transparency, e.g., to account for safety concerns or quality control. 

This led to a field of research considering the explainability of such black-box models. 
This not only concerns providing explanations for specific model decisions, but also extends to providing general interpretability of the model. 
In the following, we provide an overview of approaches for interpretable models via the taxonomy of Li et al.~\cite{Li2022}, classifying interpretation algorithms according to three characteristics: 

The first characteristic is the representation of the interpretation.
It can be based on the \textit{importance} of (latent) features in relation to the final objective \cite{lundberg2017unified}. 
Alternatively, one can use the \textit{model's response} to different inputs to identify behaviors. 
Some algorithms approximate the model using an interpretable surrogate model \cite{ribeiro2016should}. 
Finally, some models show the interpretation by a \textit{dataset} of samples that show the impact of training \cite{koh2017understanding,pleiss2020identifying}.
The second characteristic is the type of the model to be interpreted.
Some approaches consider the model as a black-box \cite{pleiss2020identifying, ribeiro2016should}.
These algorithms are called \textit{model-agnostic} and can be applied to any model. 
Other approaches require specific model characteristics such as differentiability or even a specific model type \cite{koh2017understanding}.
The third characteristic considers the relation between the interpretation algorithm and the model. 
\textit{Closed-form} algorithms are applied after training while \textit{composition} algorithms can (also) be integrated into the training process.
The other two mentioned relations are \textit{dependence}, where the algorithms add operations to the model after training to output interpretable terms, and \textit{proxy}, where an interpretable proxy model is created.

Our algorithm represents the interpretation as a form of \textit{model response}, displaying the policy behavior throughout various trajectories provoked by the initial state.
We furthermore consider the model to be a black box, where our algorithm can interpret various different models, provided any action selection probability.
The type of model is not relevant to our approach, making our approach model-agnostic. 
Furthermore, we propose a closed-form approach to be applied after training. 

\subsection{Evolutionary Optimization}
To optimize initial states that cause diverse demonstrations we use evolutionary optimization. 
Formally, we consider a population-based evolutionary optimization process $\mathcal{E} = \langle\mathbb{X}, \mathcal{F},\mathbb{P}_0, E, \rangle$, with populations $\mathbb{P} = \{\tau_i\}_{0\leq i \leq p}$ of size $p$, where the initial population $\mathbb{P}_0$ is chosen randomly, state space $\mathbb{X}$ with $\mathbb{P} \in \mathbb{N}^{\mathbb{X}}$, a fitness function $\mathcal{F}:\mathbb{X} \to \mathbb{R}$, and the evolution step function 
$$E(\mathbb{P}_t,\mathcal{F})=\mathbb{P}_{t+1}=\sigma_{p}\big(\mathbb{P}_t \setplus \textit{mutants}(\mathbb{P}_t)\setplus \textit{children}(\mathbb{P}_t)),$$
with a (non-deterministic) selection function $\sigma_n: \mps{\mathbb{X}} \to \mps{\mathbb{X}}$ that returns $n \in \mathbb{N}$ individuals and could depend on $\mathcal{F}$,
a function $\textit{mutants} = \{\texttt{mutation}(x) \; : \; x \sim \sigma_{\lceil p\cdot p_m\rceil}\}$ for some operator $\texttt{mutation} : \mathbb{X} \to \mathbb{X}$
and a function $\textit{children} = \{\texttt{crossover}(x_1, x_2) \; : \; x_1,x_2 \sim \sigma_{\lceil p\cdot p_c\rceil}\}$ for some operator $\texttt{crossover} : \mathbb{X} \times \mathbb{X} \to \mathbb{X}$, with mutation and crossover probabilities (rates) $p_m$ and $p_c$. Note that we use multi-set notation to handle duplicate individuals in a mathematically accurate way, where $\mathbb{N}^\mathbb{X}$ is the space of multi-sets over elements of $\mathbb{X}$ and $\setplus$ adds two multi-sets. \cite{fogel2006evolutionary}

Individuals $\mathcal{I}\in\mathbb{P}$ are defined by their inherent features (\textit{genotype}), in which we encode  an initial state $s_0$ sampled from the initial state distribution $\mu$ defined by the MDP. 
Their individual fitness is calculated based on their resulting appearance (\textit{phenotype}), i.e., the demonstration trajectory $\tau$ generated by executing a given policy $\pi$ in the given environment starting from $s_0$.
The individual state is typically represented as a binary encoding. 
This allows for the implementation of $\texttt{mutation}$ as a simple \textit{bit-flip} operation and $\texttt{crossover}$ as a \textit{single-point crossover} to recombine the state of two parents.
To foster parents with a higher fitness, \textit{tournament selection} is commonly applied within function $\sigma$. 
While evolutionary algorithms are usually used to search for one single best individual, we are interested in the entire population of individuals \cite{ishibuchi2008evolutionary,neumann2019evolutionary}. 
By deploying a fitness function that promotes diversity among trajectories, we can see the different strategies an agent follows in different situations.
Generally, all measures of the diversity of an individual $\mathcal{I}$ in a population $\mathbb{P}$ are related to the pairwise distance between individuals in $\mathbb{P}$ as measured by a suitable norm (e.g., Euclidean for real-valued representations, Hamming for symbolic representations) \cite{wineberg2003underlying}. 
Therefore, the individual diversity w.r.t. the population can be estimated by 
$\mathcal{D}(\mathcal{I},\mathbb{P}) = \frac{1}{p} \cdot \sum_{\mathcal{I}'\in\mathbb{P} } |\mathcal{I}'- \mathcal{I}|$ \cite{gabor2018inheritance}.

\section{Related Work}

\paragraph{\textbf{Evolutionary RL}}
Among various optimization-related applications, evolutionary algorithms (EAs) have also been applied to RL, primarily, to foster their diverse exploration capabilities. 
Therefore, EAs have been used mainly for the training of a policy, like by Kadka and Tumer~\cite{DBLP:journals/corr/abs-1805-07917}, where an actor-critic network is trained by optimizing a population of actor networks and keeping collected data and weights to update the policy. 
In regular intervals, the policies' weights are updated by the population of actors. 
This approach enables the use of a qualitative and diverse set of experiences.
Similarly, \citet{DBLP:populationbasedRL} optimize a population of policies to obtain a trained yet highly diverse population. 
In addition, they included an option to adapt the degree of diversity during training. 
They measure diversity using task-agnostic behavioral embeddings.
\citet{wu2023qualitysimilar} train several policies in parallel and optimize them at different levels of quality by looking at the task-specific diversity. 
This approach is especially useful if one is interested in a specific kind of diversity.
\citet{wang2022evolutionary} cluster a population of policies by behavior. 
From each cluster, the policy with the highest quality is selected. 
However, the behavior characterization here is highly use-case dependent since it is restricted on the final state of the agent.
All of those approaches have one thing in common: They try to find a good compromise between quality and diversity. 
This shows that quality is not the only factor to be considered when evaluating a policy. 
It is equally important to regard its diversity in order not to overfit the policy.
In our case, however, we view diversity as being strongly related to interpretability. 
The more diverse the behavior of an agent, the harder it is to interpret its policy. 
Different behaviors of an agent show different aspects of its policy.
However, this is not the usual approach to interpretability.

\paragraph{\textbf{Explainable RL}}

There are several approaches to the \textit{interpretability and explainability of RL} (XRL), which are surveyed by \citet{HEUILLET2021106685} and \citet{RLinterpretsurvey}.
Similar to general explainability approaches previously introduced, RL interpretation algorithms can be divided into different categories. 
One central aspect is the scope of the interpretation algorithm, reflecting either local decisions or the global strategy. 
Further distinction is drawn between post-hoc methods, which keep the original model while the explanation is given via a separate model, and intrinsic methods, which can be models that can replace the original model and represent a more explainable surrogate. 
Combinations of both are also possible. 
Furthermore, XRL algorithms can be applied before, during, or after training.
Finally, XRL algorithms can be classified according to their type of explanation.
The most common types are textual explanations, image explanations, a collection of states or state-action pairs, or an explanation through rules.
%
We approach XRL by generating a collection of demonstration trajectories from a given policy interacting with an environment to show diverse behavior.
We thereby strive for a scope that includes the global inherent strategy. 
Furthermore, we optimize the diversity of those demonstrations using an evolutionary process, which can be considered a post-hoc method.
Specifically, REACT does not require a particular policy specification and therefore does not need to be integrated prior to or during training.
Similarly, \textit{HIGHLIGHTS} creates a policy summary, which contains a fixed number of important states together with their surrounding states \cite{Amir2018HIGHLIGHTSSA}. 
The importance of states is identified by the effect of an action on that state.
The goal is to find states where a small modification of the action would strongly influence the cumulative reward. 
Therefore, the approach is mainly based on the value function rather than relying on an external optimization mechanism. 
Such states have also been referred to as \textit{critical states}, where the chosen action has a significant impact on the outcome, which can be used to interpret policies trained using maximum entropy-based RL \cite{DBLP:crtiticalstates}. 
While REACT and \textit{HIGHLIGHTS} share their global scope, we refrain from integrating the reward into the fitness to be optimized and instead use it to validate finding a set of diverse demonstrations. 
Likewise, \citet{SEQUEIRA2020103367} created \textit{interestingness elements} considering the agents' actions and states in the environment, but also at its policy, to be compiled into a summarizing video, similar to \textit{HIGHLIGHTS}. 
Interesting elements are determined by the frequency, execution certainty, transition value, and sequences, with the aim of showing a maximally diverse set of highlights. 
Interesting sequences are chosen by maximizing the minimal and maximal distance between pairs of sequences.
In contrast to both \textit{HIGHLIGHTS} and \textit{interestingness elements} we consider demonstrations of full trajectories instead of patching together possibly unrelated sequences due to their impact.
%
Interesting elements cannot just be found by looking at actions and states. 
\citet{DBLP:HuangHAD17} trained a robot via \textit{inverse RL} (IRL) to infer objectives from observed behavior and to select the most informative behavior to show to the user. 
Inferring the objective function is done via Bayesian inference.
On the contrary, \citet{DBLP:policysummarization} evaluated both IRL and \textit{imitation learning} (IL) for their quality in extracting good summaries. 
With IL, a function is learned to map states to actions without any knowledge of rewards or goals. 
Based on that function, probabilities and thus predictions are created. 
From these predictions, the summary is extracted by selecting state action pairs that minimize the loss of all unseen states. 
Experiments showed that summaries created by IL are more helpful, if the domain is not familiar.

\paragraph{\textbf{Robust RL}} 

We consider a process where a policy is trained with a single deterministic initial state and evaluated with a changing initial state to simulate edge-case behavior of the policy, allowing interpretability of the learned behavior.
Therefore, from a different perspective, we consider the robustness of a policy to out-of-distribution samples, i.e., initial states that were potentially not experienced during training, also referred to as generalization capabilities. 
If an agent is trained well, only looking at some episodes of the agent's interaction with the environment usually only shows the expected behavior, including often-occurring states. 
However, the agent's strategy also includes behavior in states that have not been encountered that often. 
We also want to show this behavior. 
The goal is to show the most diverse behavior and generate a small but informant overview of the agent's strategy.
To improve the generalization capabilities, using varying training configurations has shown to be a viable approach \cite{cobbe2020leveraging}.
Yet, we specifically chose a different training approach to showcase the methodical impact of REACT for visualizing a possibly insufficient policy in edge-case scenarios. 
Note that this work, in general, does not consider any kind of policy improvement.
We intend to assess the intricacies of a trained policy by representing its behavior in a suitable interpretable way. 
Nevertheless, the scenarios generated by the proposed architecture could be used as \textit{adversarial samples} and thus fed back into the training process to further improve the policy, similar to \citet{gabor2019scenario}.

\section{Trajectory Fitness Evaluation}

\begin{figure}\centering
    \includegraphics[width=0.8\linewidth]{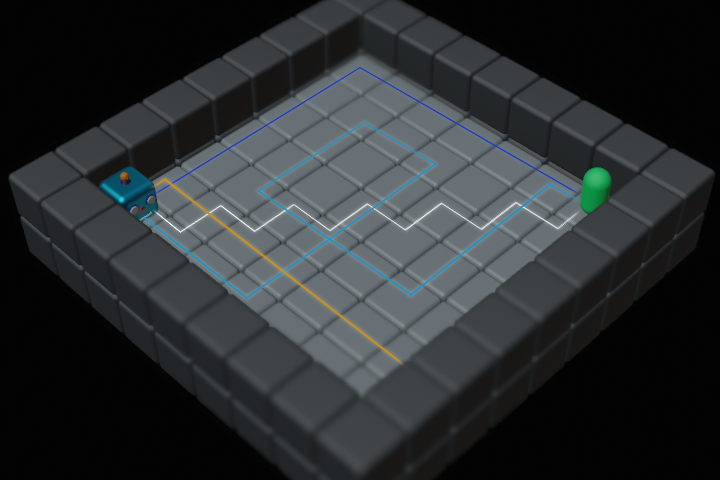}
    \caption{Joint fitness $\mathcal{F}$ elements \textit{local diversity} $\mathcal{D}_l$ (light blue), \textit{global diversity} $\mathcal{D}_g$ (blue), and \textit{certainty} $\mathcal{C}$ (orange), compared to an exemplary optimal trajectory (white).}
    \Description{Visualization of the joint fitness $\mathcal{F}$ elements \textit{local diversity} $\mathcal{D}_l$ (light blue), \textit{global diversity} $\mathcal{D}_g$ (blue), and \textit{certainty} $\mathcal{C}$ (orange), compared to an exemplary optimal trajectory (white).}
    \label{fig:joint_fitness}
\end{figure}

In the following, we discuss assessing the fitness of trajectories $\tau=\langle s_0, a_0, r_1, \dots, s_t, a_t, r_{t+1}\rangle \sim  \mathcal{P}_{\pi,s_0}$ to serve as an \textit{insightful} demonstration to interpret the inherent behavior of policy $\pi$.
Unlike the central objective of RL, we are not interested in optimizing for the best-performing individuals, but rather interested in a population of diverse demonstrations following $\pi$ from an initial state $s_0$ to be optimized. 
Therefore, we refrain from using the reward metric supplied to learn the policy and define a joint fitness metric $\mathcal{F}$ in the following.
To illustrate our deliberations, we consider a simple $9\times9$ gridworld environment as depicted in Fig.~\ref{fig:joint_fitness}.

For \textit{insightful} demonstrations, we strive to achieve high diversity. 
Considering a single trajectory, a diverse path, covering a larger fraction of the available state space (e.g., the light blue path in Fig.~\ref{fig:joint_fitness}), would be more informative regarding the behavior to be analyzed than the comparably direct path we would expect resulting from policy optimization (e.g., Fig.~\ref{fig:joint_fitness}, white path). 
Even though it might be considered less optimal w.r.t. the reward of the given environment, such behavior might depict an edge, important to assess the given policy. 
We refer to this measure as \textit{local diversity} and formalize the corresponding metric 
\begin{equation}\label{eq:local_diversity}
\mathcal{D}_l(\tau) = \frac{| \{ s \in \tau\}|}{|\{s \in \mathcal{S}\}|}.
\end{equation}
Overall, the higher the local diversity, the more divergence from the optimal path occurs, increasing the relevance of the trajectory. 
Note that, for simplicity, we only consider state disturbances of the agent position.
Therefore, any grid with $n \times m$ fields holds $|\{s \in \mathcal{S}\}|=n\cdot m$ distinct states.
Furthermore, this allows us to consider $\rho(s)$ to extract a position from state $s$.
For continuous state spaces, we replace the total number of possible states $|\{s \in \mathcal{S}\}|$ by the trajectory length $|\tau|$.

Considering multiple trajectories $\mathcal{T}$, however, neither solely disturbed (light blue) nor solely optimal (white) paths accurately reflect the behavior of $\pi$.
We therefore additionally consider a \textit{global diversity} $\mathcal{D}_g$ (blue) of trajectories $\tau \in \mathcal{T}$ formalized as
\begin{equation}\label{eq:global_diversity}
\mathcal{D}_g(\tau, \mathcal{T}) = \frac{1}{\lceil\mathcal{S}\rceil}\min_{t\in \mathcal{T}\setminus\tau}\delta(\tau,t)
\end{equation}
based on the \textit{maximum state distance} $\lceil\mathcal{S}\rceil=\max_{s, t \in \mathcal{S}}||s-t||_2$ and the  \textit{one-way distance} $\delta$ between trajectories $U$ and $V$ \cite{Lin2008OneWD}: 
\begin{equation}
\delta(U,V) =  \frac{1}{|U|+|V|}  \left(\sum_{p \in u} d(p, V) + \sum_{q \in v} d(q, U)\right),
\end{equation}
using the state-to-trajectory distance $d$:
\begin{equation}
d(p,T)=\min_{t\in T}(||s-t||_2),
\end{equation}
with the Euclidean state distance $||s-t||_2=\sqrt{(\rho(s)-\rho(t)^2}$ based on the agent position.
This accumulated two-way measure allows for comparison between trajectories of different lengths. 
Furthermore, using the $\min$ operator in Eq.~\eqref{eq:global_diversity} causes equal trajectories in $\mathcal{T}$ to be valued at $0$. 
Ultimately, even if $\mathcal{T}$ contains only optimal, yet maximally dissected behavior to reach the target, presenting such diverse demonstrations increases the overall interpretability of $\pi$. 
Note that, even though only defined for disturbances regarding the agent position, introducing further deviations, such as altering the layout, is formally not precluded. 
To calculate the global diversity, however, this might require using a different distance metric, like the Levenshtein distance, instead.

Both diversity measures implicitly cover insufficiencies and uncertainties of $\pi$ that may occur in states less prevalent during training. 
To reflect the diversity of the action decision itself, we furthermore consider the \textit{certainty}, formalized as the cumulative normalized action probability of $\tau$ given $\pi$:
\begin{equation}\label{eq:certainty}
\mathcal{C}_\pi(\tau) = \frac{1}{|\tau|}\sum_{s,a\in\tau}\pi(a|s).
\end{equation}
Counterintuitively, we are interested in trajectories with low certainties causing more diverse decisions that may result in failure to solve the intended task, such as the exemplary orange path in Fig.~\ref{fig:joint_fitness}.

Overall, we define the \textit{joint fitness}, combining \textit{global diversity} $\mathcal{D}_g$, \textit{local diversity} $\mathcal{D}_l$, and \textit{certainty} $C_\pi$ of a trajectory $\tau$ in context of a set of previously evaluated trajectories $\mathcal{T}$ as follows: 
\begin{equation}\label{eq:joint_fitness}
    \mathcal{F}(\tau,\mathcal{T}) = \mathcal{D}_g(\tau, \mathcal{T}) + \min_{t\in\mathcal{T}}\left|\left|\begin{pmatrix}\mathcal{D}_l(\tau)\\\mathcal{C}_\pi(\tau)\end{pmatrix}-\begin{pmatrix}\mathcal{D}_l(t)\\\mathcal{C}_\pi(t)\end{pmatrix} \right|\right|_2.
\end{equation}
To reflect the $\tau$-specific metrics of local diversity and certainty in relation to the set of trajectories $\mathcal{T}$, considered for calculating the \textit{global diversity}, we consider these measures only regarding their minimum distance between $\tau$ and $\mathcal{T}$.
Furthermore, we combine them into a vector to foster diverse and uncertain trajectories.
As we defined all components of the joint fitness to be normalized, we furthermore do not introduce additional parameterizations to balance their impact.
Preliminary studies confirmed this approach.

\section{REACT}
\begin{figure}[t]\centering  
\includegraphics[width=0.9\linewidth]{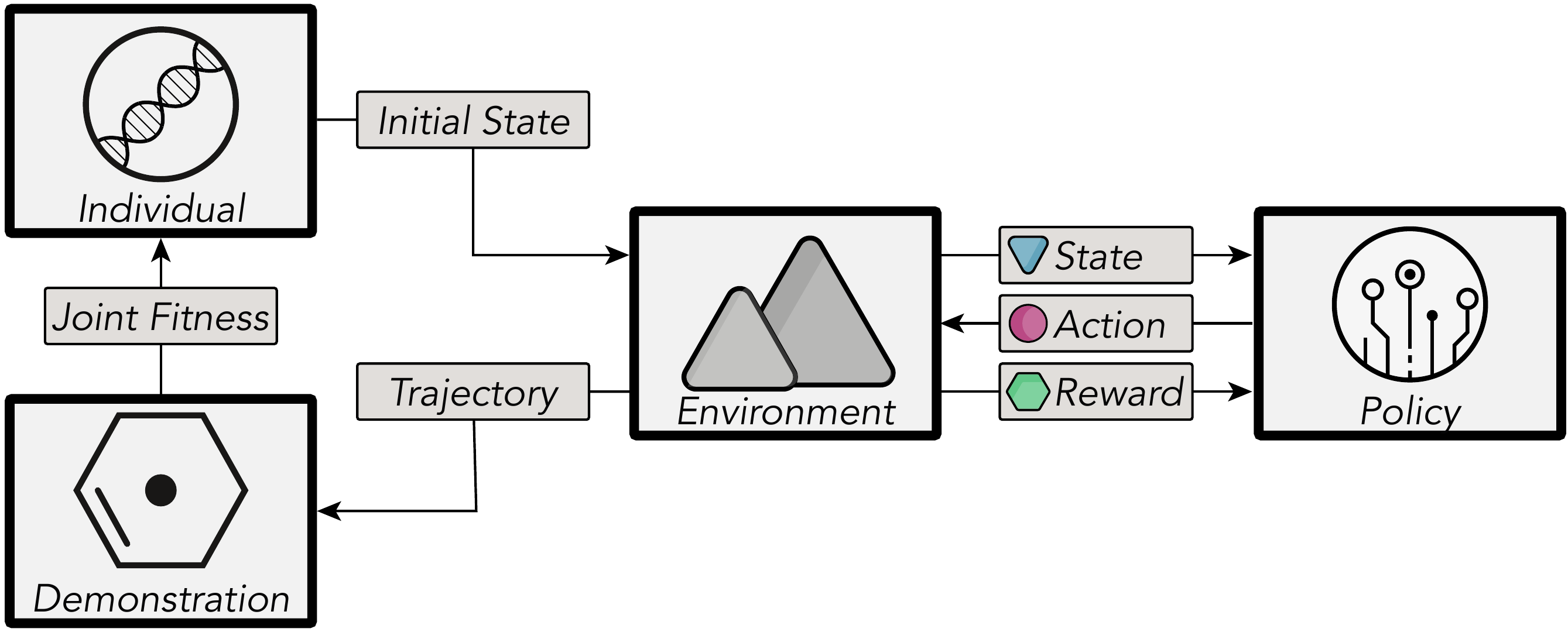}
\caption{REACT Architecture}\label{fig:react}
\Description{REACT Architecture}
\end{figure}

To optimize a pool of demonstrations to serve for interpreting a given policy using the previously defined fitness, we propose \textit{revealing evolutionary action consequence trajectories} (REACT) to optimize a population of initial states causing diverse demonstrations. 
By showing not only the optimistic optimal behavior, we aim to increase the traceability of the learned behavior and ultimately trust in the black-box policy model. 
In contrast to most evolutionary approaches, we are interested in the whole population, not just the single best-performing individual.
The overall architecture is depicted in Fig.~\ref{fig:react} and outlined in Alg.~\ref{alg:react}.

\begin{algorithm}[ht]
\caption{REACT\protect\footnotemark}\label{alg:react}
\begin{algorithmic}[1]
\REQUIRE{policy $\pi$, environment transition probabilities $\mathcal{P}$, initial state distribution $\mu$, population size $n$, generations $g$, crossover and mutation probabilities $p_c, p_m$}
\STATE{$\mathbb{P} \gets \langle s_0 \sim \mu\rangle_n$ \quad \textit{$\triangleright$ Generate initial population of size $n$}}
\STATE{$\mathcal{T} \gets \emptyset$ \hspace{3.5em} \textit{$\triangleright$ Initialize empty demonstrations}}
\FOR{individual $\mathcal{I} \in \mathbb{P}$}\label{alg:ln:Fs}
\STATE{$\tau \sim \mathcal{P}_{\pi, s_0}$\quad \textit{$\triangleright$ Sample trajectory $\tau_\mathcal{I}$ from initial state $s_0$}}
\STATE{$\mathcal{F}_\mathcal{I}(\tau,\mathcal{T})$\quad\textit{$\triangleright$ Calculate Fitness of $\mathcal{I}$ w.r.t. to phenotype $\tau$ and \\ \hspace{4.8em} previous demonstrations $\mathcal{T}$ according to Eq.~\eqref{eq:joint_fitness}}}
\STATE{$\mathcal{T}\gets\mathcal{T}\cup\tau$\quad \textit{$\triangleright$ Update demonstrations $\mathcal{T}$}}
\ENDFOR\label{alg:ln:Fe}
\FORALL{$g$ generations}
\STATE{$\triangleright$\textit{ Generate offspring from crossover with tournament selection based on individual fitness and $p_c$ and mutation based on $p_m$}}
\STATE{$\mathbb{O}\gets \texttt{crossover}(\mathbb{P}, \mathcal{F}, p_c)\setplus\texttt{mutation}(\mathbb{P}, p_m)$}
\STATE{\textit{$\triangleright$ Calculate offspring fitness according to lines~\ref{alg:ln:Fs}--\ref{alg:ln:Fe}}}
\STATE{\textit{$\triangleright$ Select $n$ best individuals for next generation from population and generated offspring according to their fitness}}
\STATE{$\mathbb{P}\gets\texttt{migration}(\mathbb{P}\setplus\mathbb{O}, \mathcal{F}, n)$}
\STATE{$\mathcal{T} \gets \mathcal{T} \setminus \{\tau_{\mathcal{I}} \; \vert \; \mathcal{I} \notin \mathbb{P}\}$\quad\textit{$\triangleright$ Remove extinct demonstrations}}
\ENDFOR
\RETURN $\mathbb{P}$
\end{algorithmic}
\end{algorithm}

To form the initial population $\mathbb{P}$ of size $n$, individuals encoded by the initial state $s_0$ are generated from $\mu$ given by the MDP of the given environment. 
Invalid individuals that cannot generate any demonstration are disregarded.
As we only introduce disturbances related to the initial position of the agent, the initial state $s_0$ can be encoded by the initial position of the agent. 
To account for evaluation environments comprising different-sized 2d-discrete and 3d-continuous state spaces, we opt for a universal multi-dimensional bit encoding of length $6$ with inverse normalization to ensure precise reconstruction of the intended position. 
For further details, please refer to the appendix. 

To evaluate the individuals' fitness, trajectories $\tau$ are sampled from the environment, starting from their individual initial state, following policy $\pi$.
For improved comparability, we furthermore remove duplicate consecutive states from the demonstrations.
These demonstrations constitute the individuals' phenotype directly affecting their fitness to serve as a viable and diverse representation of the given model. 
As we are interested in a pool of demonstrations over a single \textit{best} demonstration, the individual fitness is calculated based on the set of previous demonstrations $\mathcal{T}$ in addition to the individual trajectory $\tau$, following Eq.~\eqref{eq:joint_fitness}.
Note that, even though we generate interactions with the environment, we do not consider the generated experience to further improve the policy at hand. 
Nevertheless, the proposed architecture could serve as an automated adversarial curriculum to generate scenarios in which the policy could be further improved.

After the evaluation of the first generation, the best individuals are selected via tournament selection to create new individuals via recombination.
The recombination operator is executed with the recombination probability $p_c \in [0,1]$ defined beforehand.
To generate the offspring, we use single-point crossover. 
The new individuals are then added to the population.
Then, a mutation operator with mutation probability $p_m \in [0,1]$ is applied to random individuals from the original population.
Mutation is implemented by a single bit-flip of one random bit in the individual's encoding. 
As we are interested in the whole population, we keep the individual before mutation and add the mutated individual to the population to keep the evolution elitist.
After evaluating the newly generated offspring, as described above, one after the other, the population is reduced to the intended size $n$ by removing the individuals with the lowest fitness value along with their generated demonstrations. 
The described procedure is repeated for a fixed number of $g$ generations.

\paragraph{\textbf{Hyperparameters}}
The most important hyperparameter to consider is the population size $p$.
It influences the effectiveness of the evolutionary process and it determines the number of demonstrations generated for interpreting the policy. 
To suit human needs, $p$ should be comprehensibly small and sufficiently diverse.
Preliminary experiments suggest that a population size of $p=10$ is a reasonable compromise.
Larger populations can be used if only the best $n$ individuals are considered to demonstrate the behavior of the policy.  
For experimental details, please refer to the appendix. 
Furthermore, if not stated otherwise, we optimize the population of demonstrations over 40 iterations (generations). 
As our central goal is to diversify the population throughout optimization, we use a reasonably high \textit{crossover probability} $p_c=0.75$ combined with a high \textit{mutation probability} $p_m=0.5$. 
In combination with the chosen binary state encoding of length $6$, representing the agent's initial position, these hyperparameter settings cause the generation of offspring starting at further distances. 

\begin{figure*}\centering
\subfloat[FlatGrid11\label{fig:FlatGrid11}]{\includegraphics[width=0.2\linewidth]{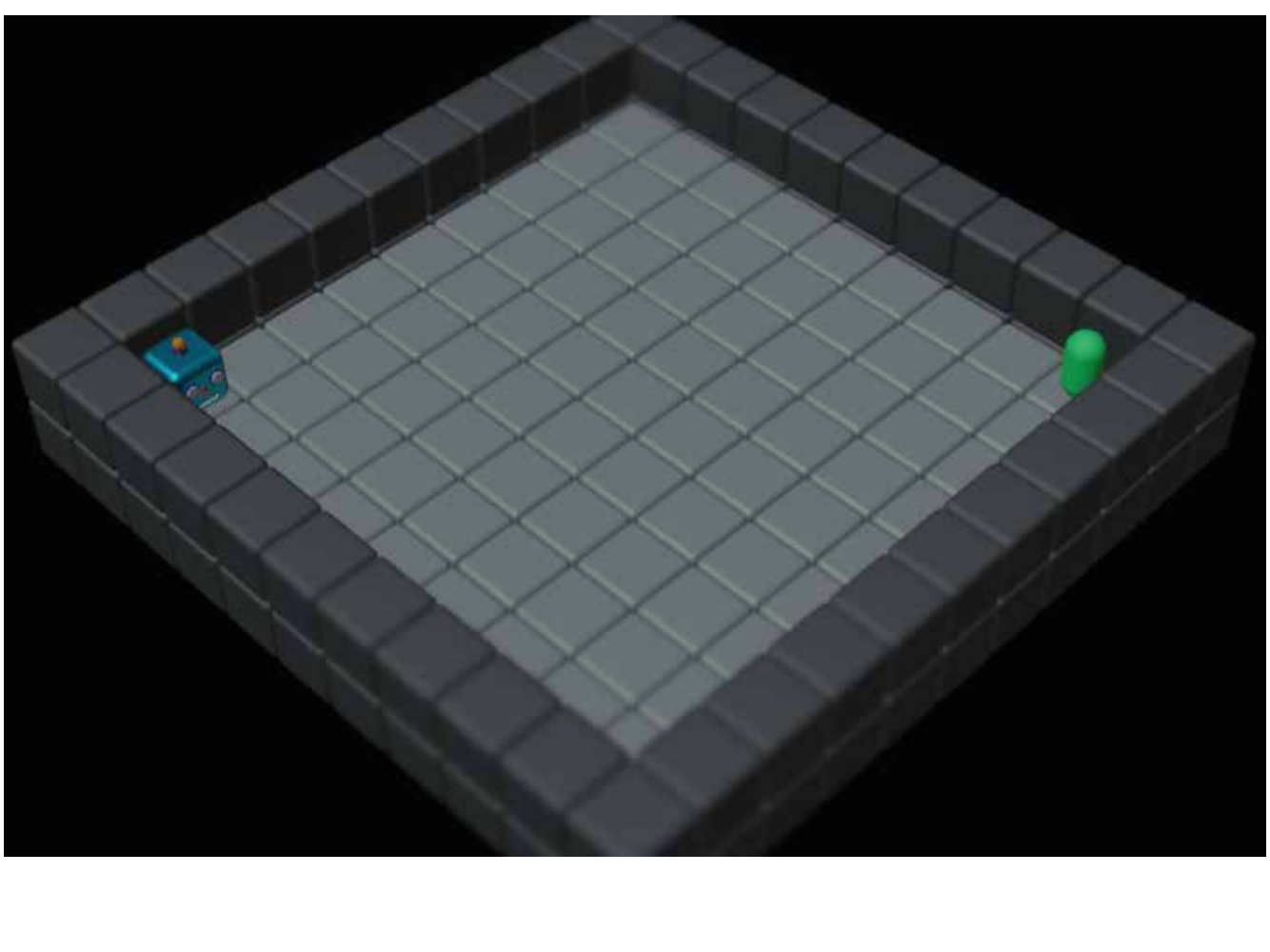}}
\subfloat[Final Return\label{fig:FlatGrid-Return}]{\includegraphics[width=0.2\linewidth]{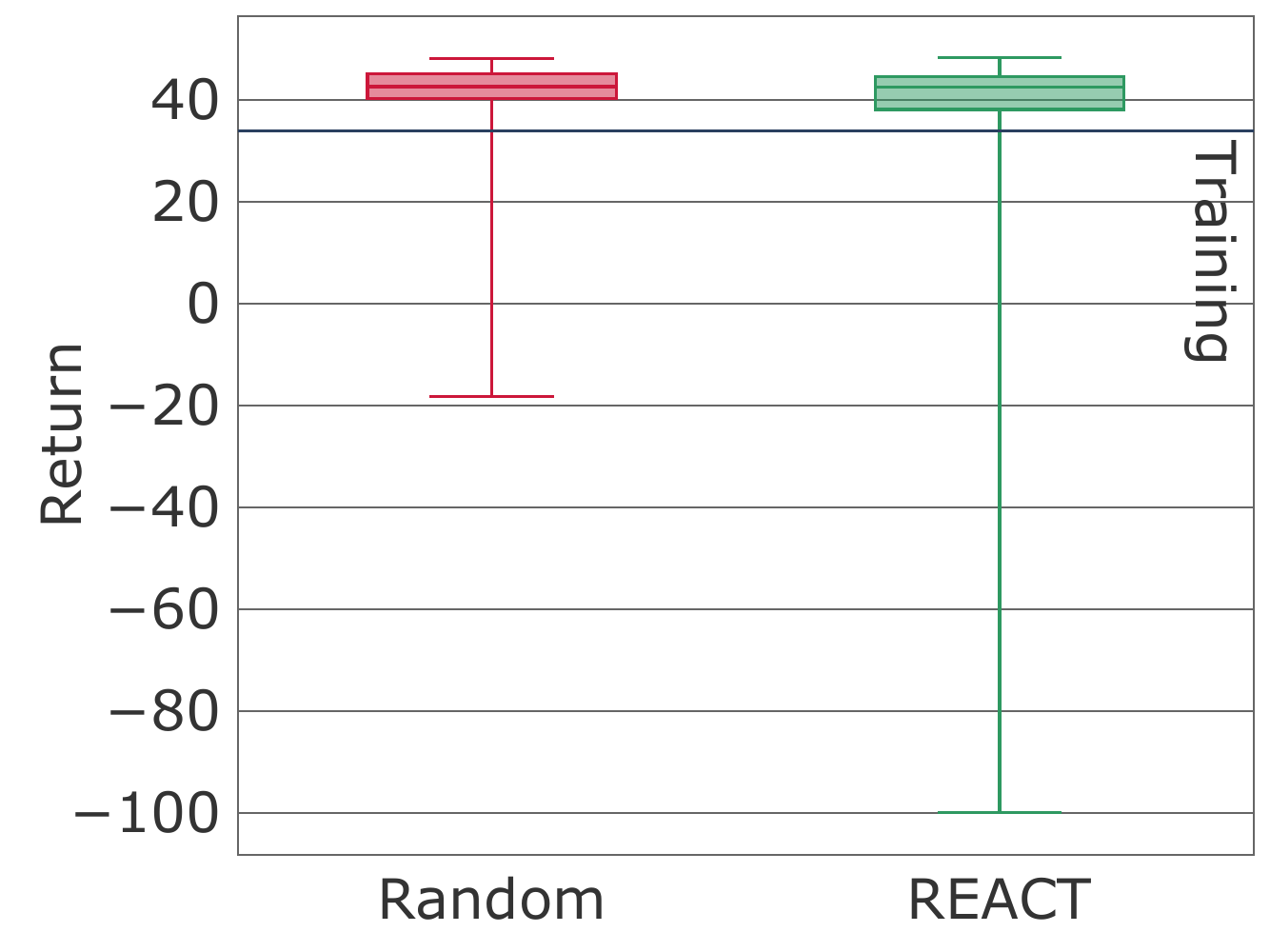}}
\subfloat[Final Length\label{fig:FlatGrid-Length}]{\includegraphics[width=0.2\linewidth]{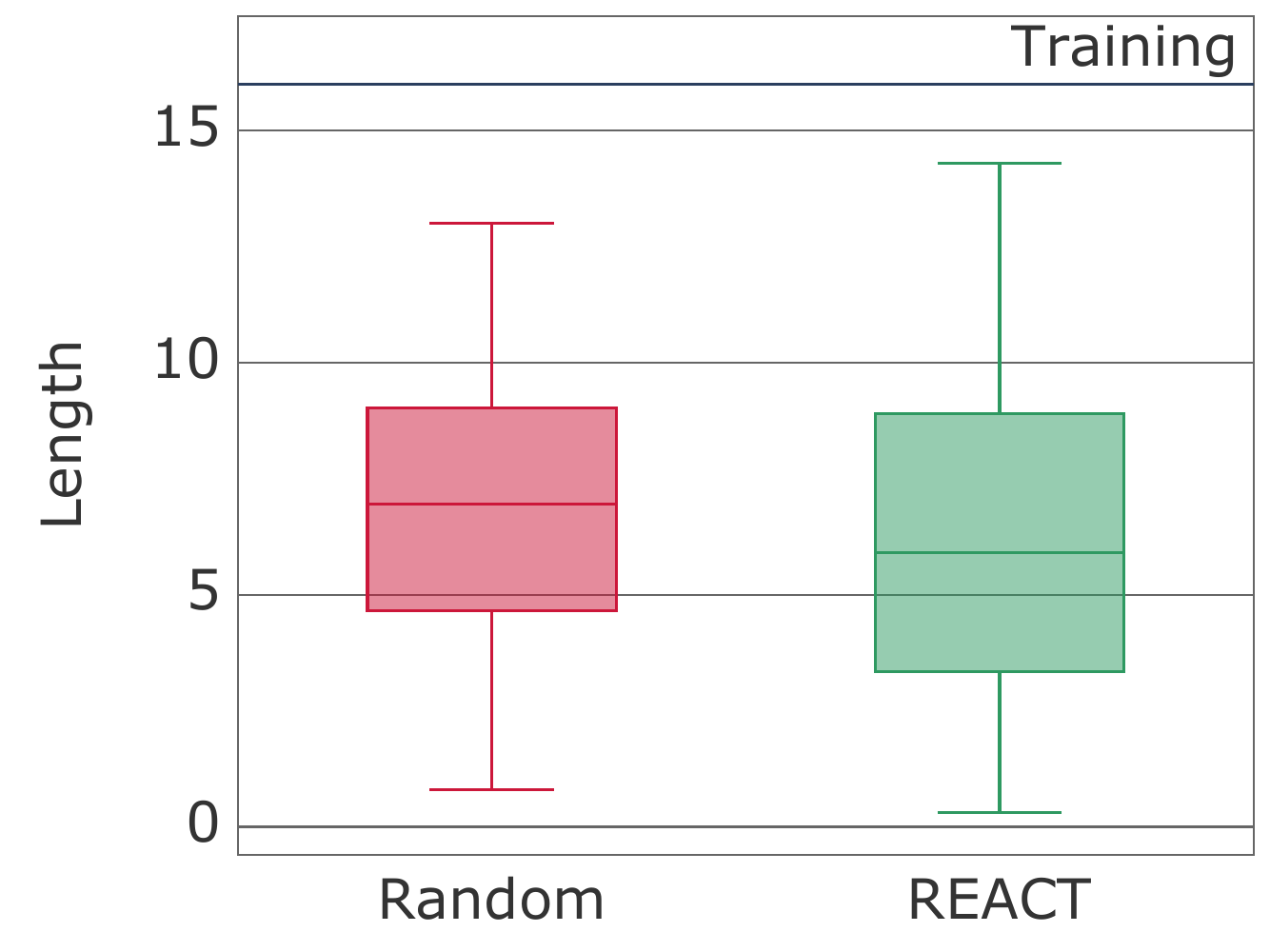}}
\subfloat[Random $\mathcal{T}$\label{fig:RandomT-FlatGrid}]{\includegraphics[width=0.2\linewidth]{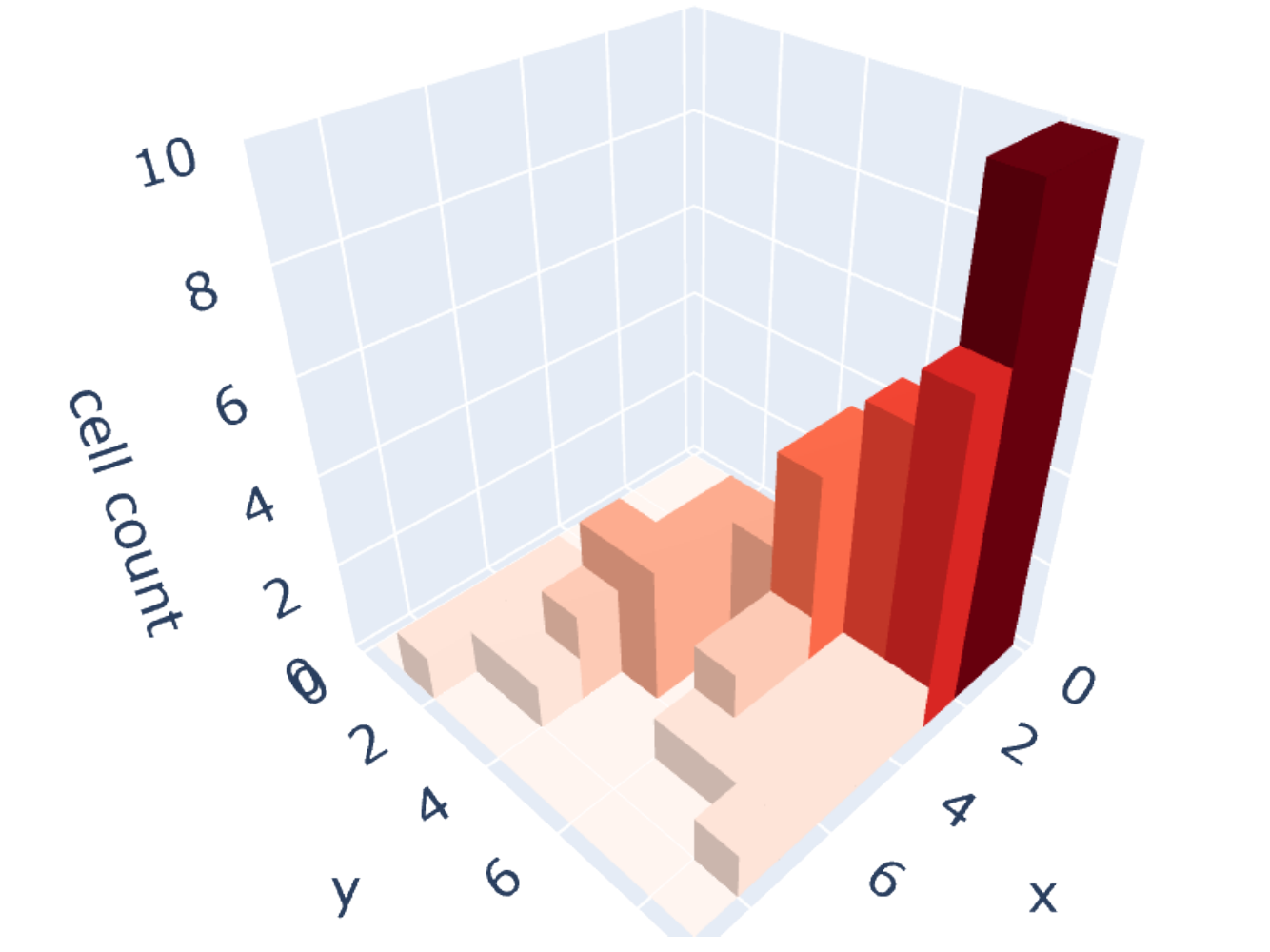}}
\subfloat[REACT $\mathcal{T}$\label{fig:ReactT-FlatGrid}]{\includegraphics[width=0.2\linewidth]{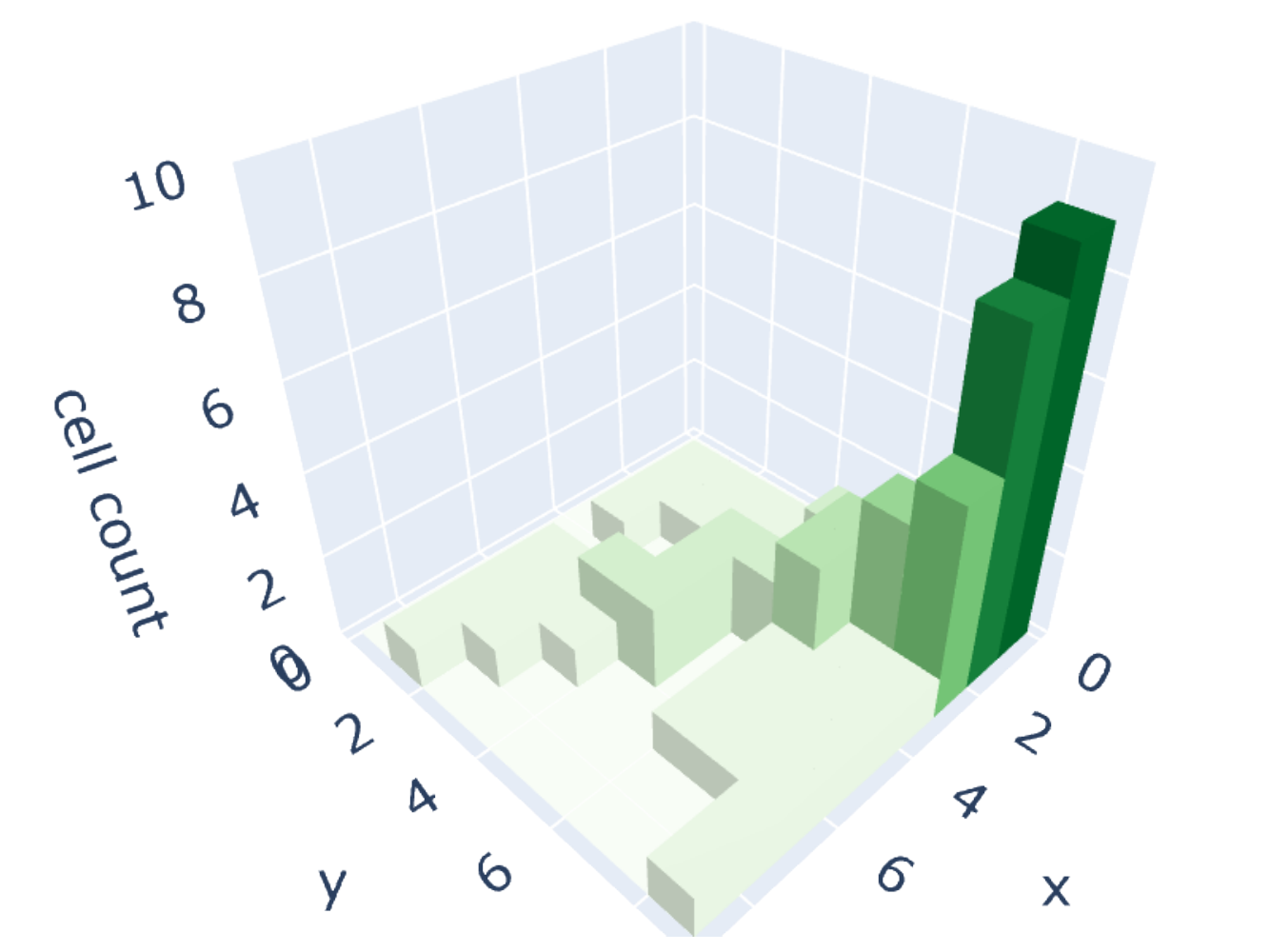}}
\caption{REACT Evaluation: {\normalfont Comparison of the Final Return~\subref{fig:FlatGrid-Return} and Length~\subref{fig:FlatGrid-Length} of Random~\subref{fig:RandomT-FlatGrid} and REACT~\subref{fig:ReactT-FlatGrid} demonstrations of a PPO policy trained for 35k steps in the FlatGrid11~\subref{fig:FlatGrid11}. The training performance in the unaltered environment is displayed by a solid line.
Overall, the plots convey the increased diversity and even distribution of REACT-generated demonstrations over random or static initial states.}} \label{fig:Eval-FlatGrid}
\Description{REACT Evaluation}
\end{figure*}

\section{Validation}

\paragraph{\textbf{Setup}}
To validate the proposed architecture, we use a simple fully observable discrete \textit{FlatGrid11} environment with $11\times11$ fields shown in Fig.~\ref{fig:FlatGrid11}\cite{Altmann_hyphi_gym}. 
The goal of the policy is to reach the target state (rewarded $+50$), where there are neither holes nor obstacles that could disrupt the agent's path.
To encourage choosing the shortest path, a step cost of $-1$ is applied. 
Episodes are terminated upon reaching the target state or after 100 steps. 
%
We use PPO \cite{schulman2017proximal} with default parameters \cite{stable-baselines3} to train a policy which we can than evaluate with REACT.
To show diverse behavior, we intentionally terminated training early (after 35k steps), just after the agent confidently reaching the target.
Using such an imperfect policy has a higher probability that the agent has not yet explored the entire environment. 
For evaluation purposes, we also want it to display behavior that leads the agent to not reach its goal.
%
To increase the significance of the experimental results presented, the following results are averaged over ten random seeds to optimize the demonstrations based on a single, previously trained policy. 

\paragraph{\textbf{Metrics}}

To provide an intuition over the resulting demonstrations $\mathcal{T}$, we summarize them in a single \textit{3D histogram}, displaying the state-frequency of all grid cells.
Compared to showing the discrete paths (cf. Fig.~\ref{fig:joint_fitness}), this has the advantage of allowing to visualize results over multiple optimization runs, without diminishing the depiction of the demonstration diversity by averaging them. 
%
Since viewing the behavior diversity of the final demonstrations is very subjective, we additionally consider the metrics \textit{final return} and \textit{final (trajectory) length}. 
Both metrics are crucial when training the optimal policy (maximizing the return while minimizing the solution length) and do not influence the optimized fitness function.
Note that the final length is usually below the episode length, as we remove recurring states to improve comparability.
However, we are not interested in the minimum or maximum of the returns or lengths, but instead in the range of the metrics and how uniformly the individuals are spread across different returns and trajectory lengths. 
Therefore, we use box-plots to visualize our results, where a bigger range between the whiskers promises greater diversity, and larger boxes indicate an even distribution. 
To provide a baseline, we also report the deterministic policy performance in the unaltered training environment, which is often used to validate learned behavior. 
Furthermore, we compare REACT to a random search approach, implemented as the initial population $\mathbb{P}_0$ before applying the evolutionary process.
This \textit{Random} approach could be considered most closely related to comparable interpretability approaches, altering the environment without optimization while maintaining comparability to REACT.

\paragraph{\textbf{Results}}
Fig.~\ref{fig:Eval-FlatGrid} shows the evaluation results. 
The trained policy reaches a return of $34$ with a trajectory length of $16$, as shown in Figs.~\ref{fig:FlatGrid-Return} and \subref{fig:FlatGrid-Length} respectively. 
Using a random pool of initial states increases the encountered return, while reducing the trajectory length of the resulting demonstrations by moving the initial state closer to the target. 
Yet, random demonstrations still mostly yield behavior in the upper reward region. 
REACT manages to further diversify the pool of demonstrations, more evenly covering a larger region of final returns.
The final lengths also show this tendency, while decreasing the overall lengths compared to the initial training state, causing a more compact representation of the learned behavior. 
Note that this outcome is further fostered by removing recurrent states from the demonstration. 
Analyzing the resulting demonstrations from a single population, shown in Figs.~\ref{fig:RandomT-FlatGrid} and \subref{fig:ReactT-FlatGrid}, reveals two further insights:
Overall, most trajectories successfully reach the target, shown by the highest occurrence of the target state, indicating a successfully trained policy that is robust to the introduced state disturbances. 
Yet, REACT produces more diverse trajectories, distributed over farther states, where some of the states even result in failure of the policy to navigate to the target. 

\begin{figure}[b]\centering
\includegraphics[width=\linewidth]{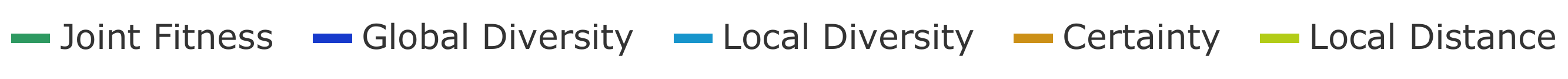}\\
\subfloat[Population Analysis\label{fig:FlatGrid-Population}]{\includegraphics[width=0.5\linewidth]{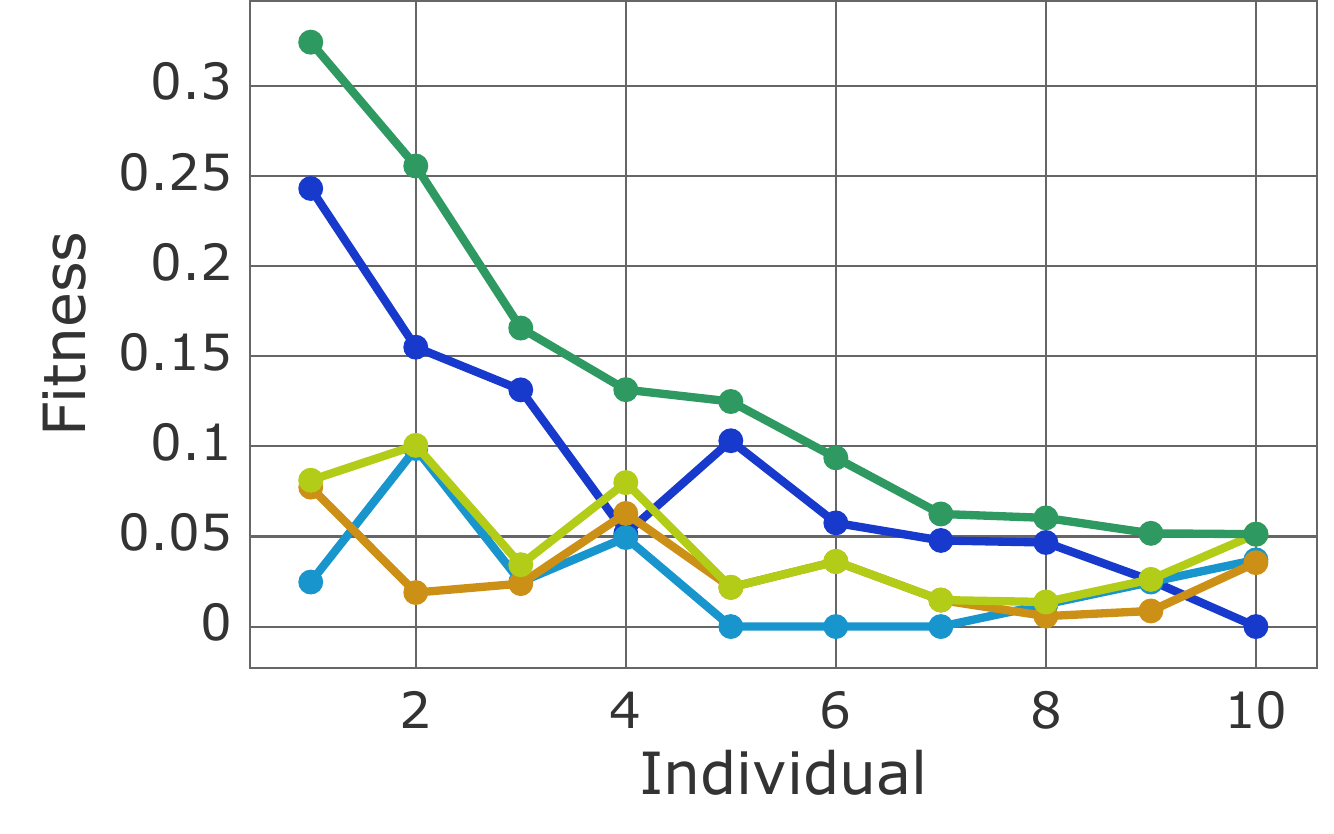}}
\subfloat[Generation Analysis\label{fig:FlatGrid-Generations}]{\includegraphics[width=0.5\linewidth]{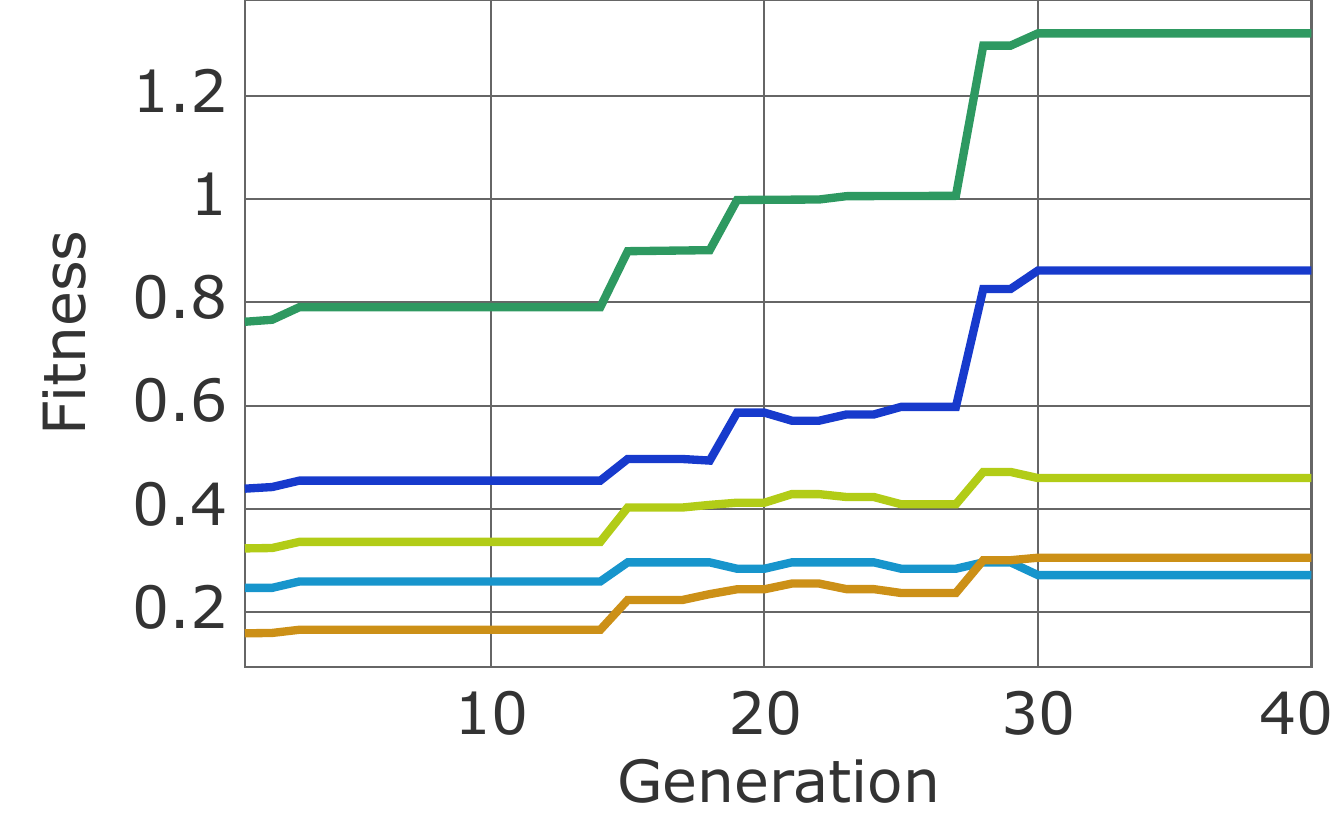}}
\caption{FlatGrid \textit{JointFitness} Analysis\label{fig:FlatGrid-Fitness}}
\Description{Joint Fitness Anaylsis}
\end{figure}

\paragraph{\textbf{Fitness Impact}} 
Besides yielding diverse demonstration, we also want to ensure the appropriateness of the proposed \textit{joint fitness}. 
Fig.~\ref{fig:FlatGrid-Fitness} therefore provides an additional in-depth analysis of the impact of the fitness components both across the single last population of $10$ individuals \subref{fig:FlatGrid-Population} and throughout the $40$ optimization generations \subref{fig:FlatGrid-Generations}.
%
\begin{figure*}\centering
\subfloat[HoleyGrid11\label{fig:HoelyGrid11}]{\includegraphics[width=0.2\linewidth]{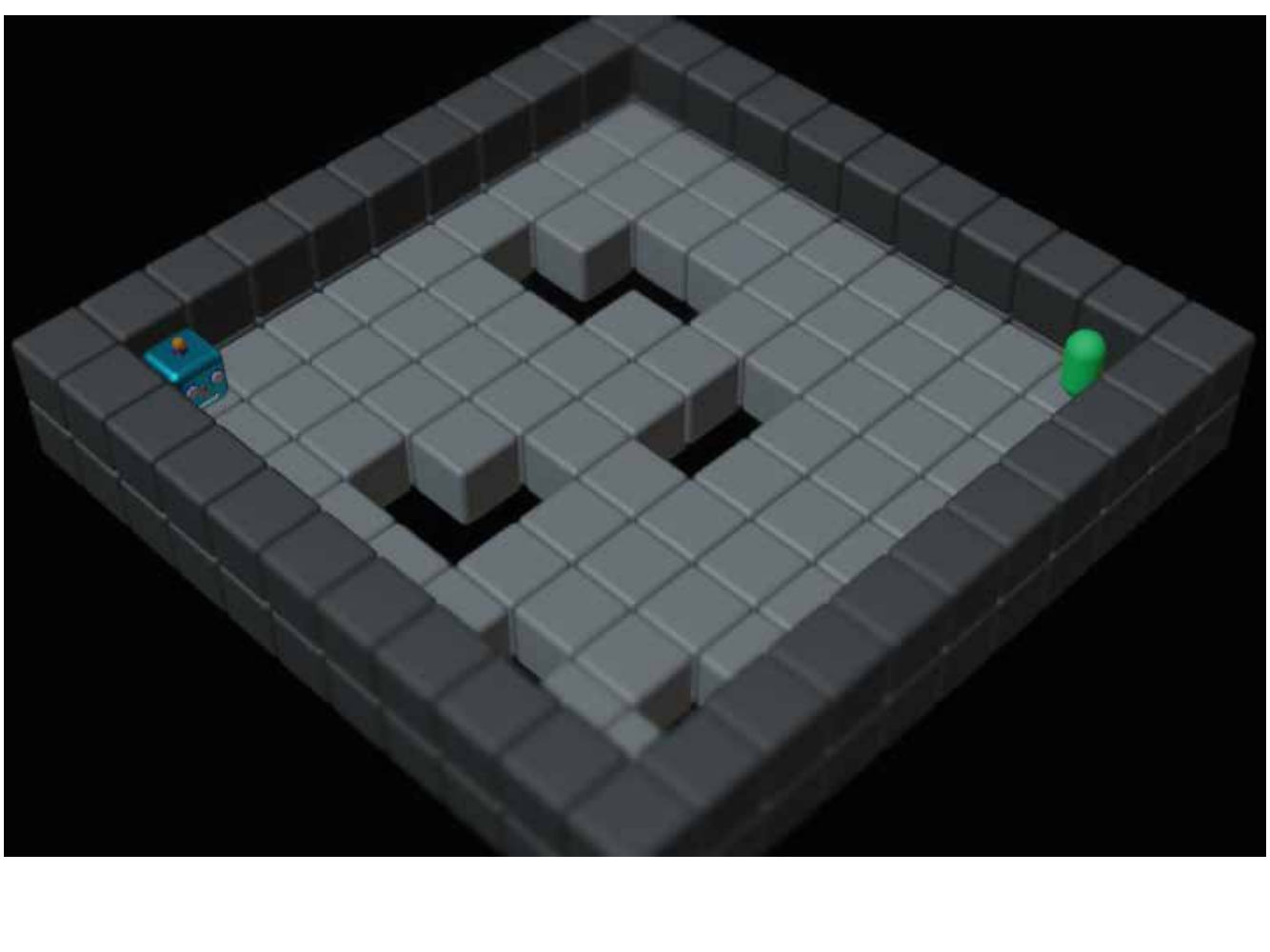}}
\subfloat[Final Return\label{fig:HoleyGrid-Return}]{\includegraphics[width=0.2\linewidth]{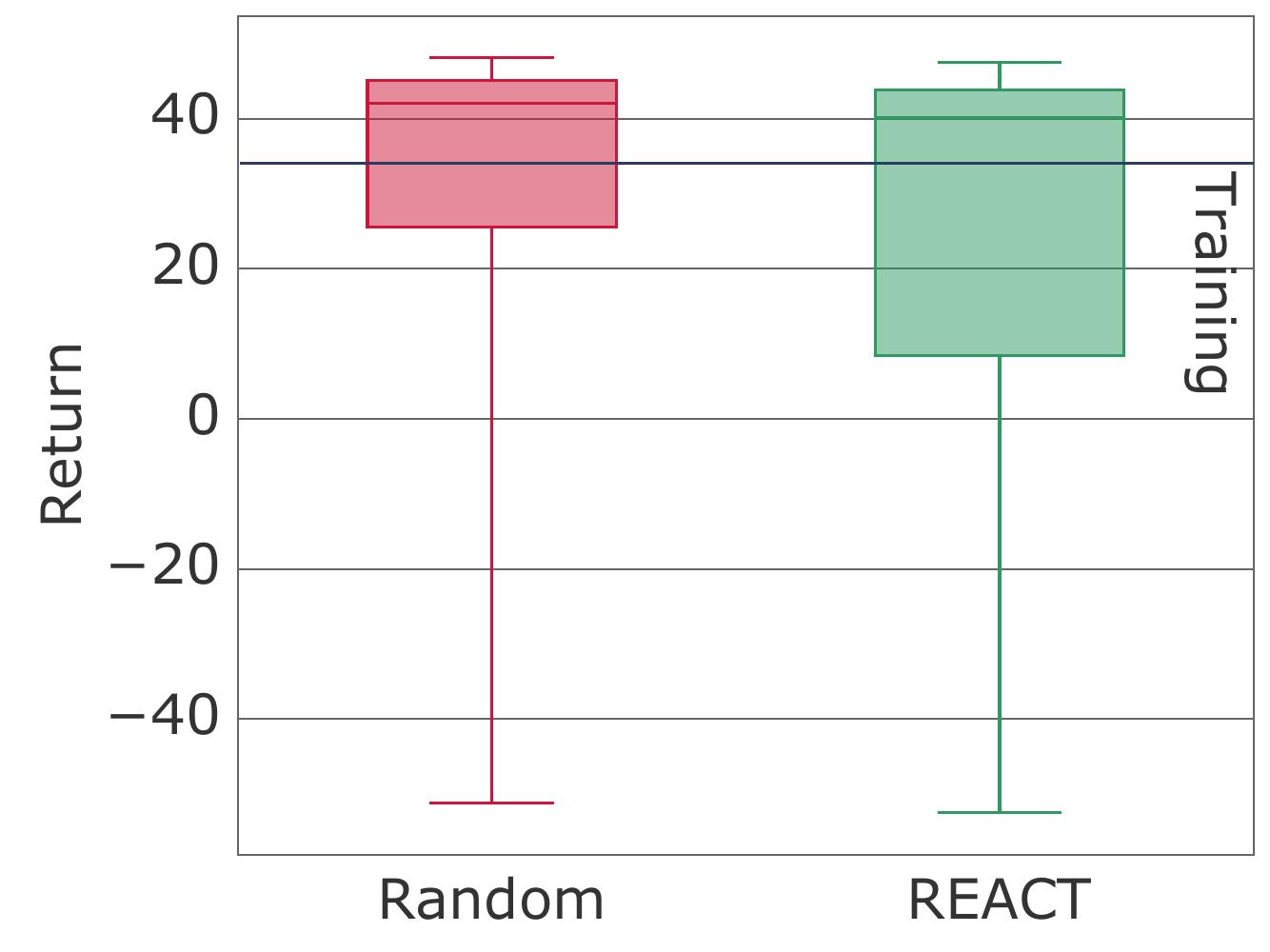}}
\subfloat[Final Length\label{fig:HoleyGrid-Length}]{\includegraphics[width=0.2\linewidth]{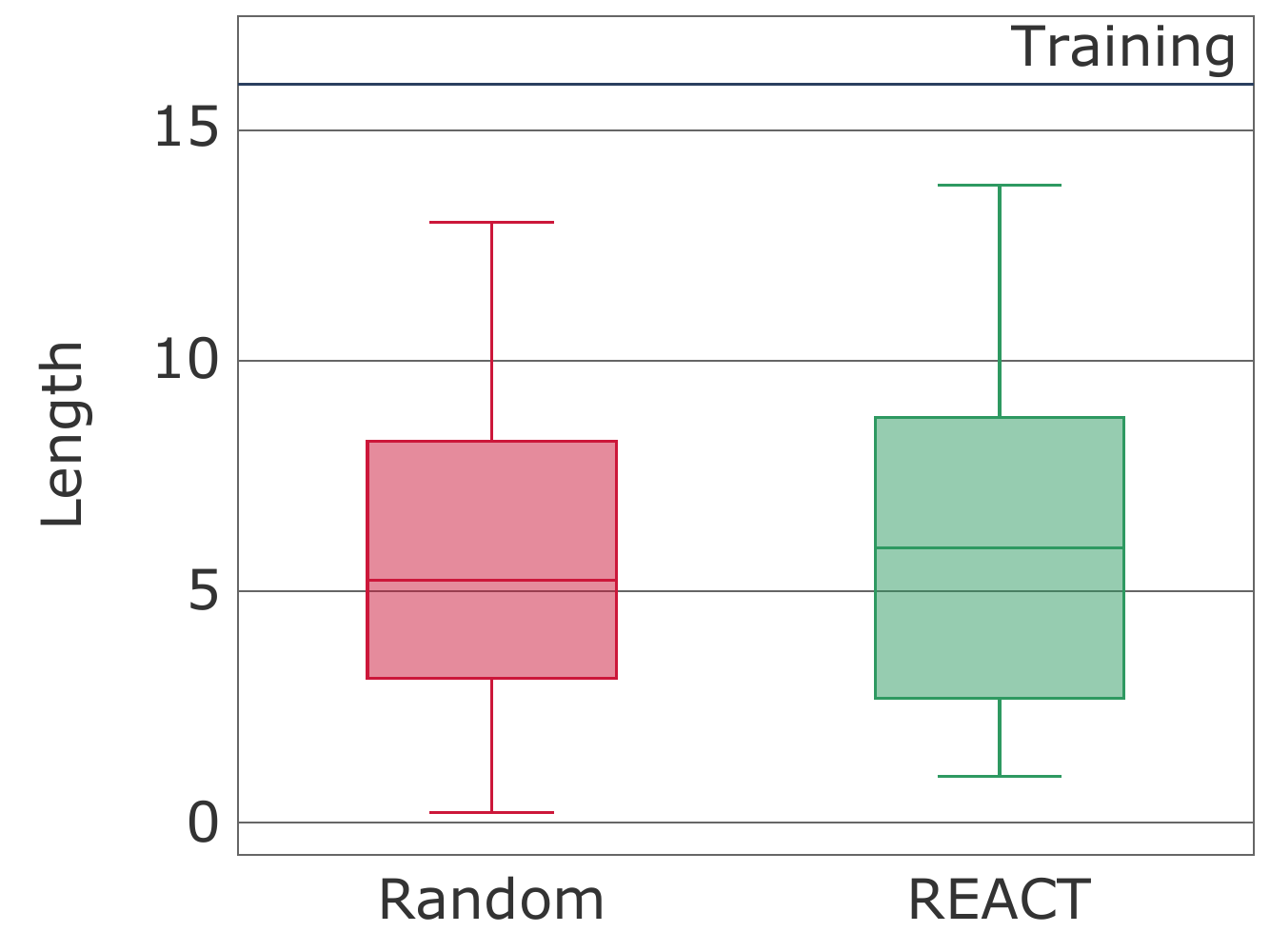}}
\subfloat[Random $\mathcal{T}$\label{fig:RandomT-HoleyGrid}]{\includegraphics[width=0.2\linewidth]{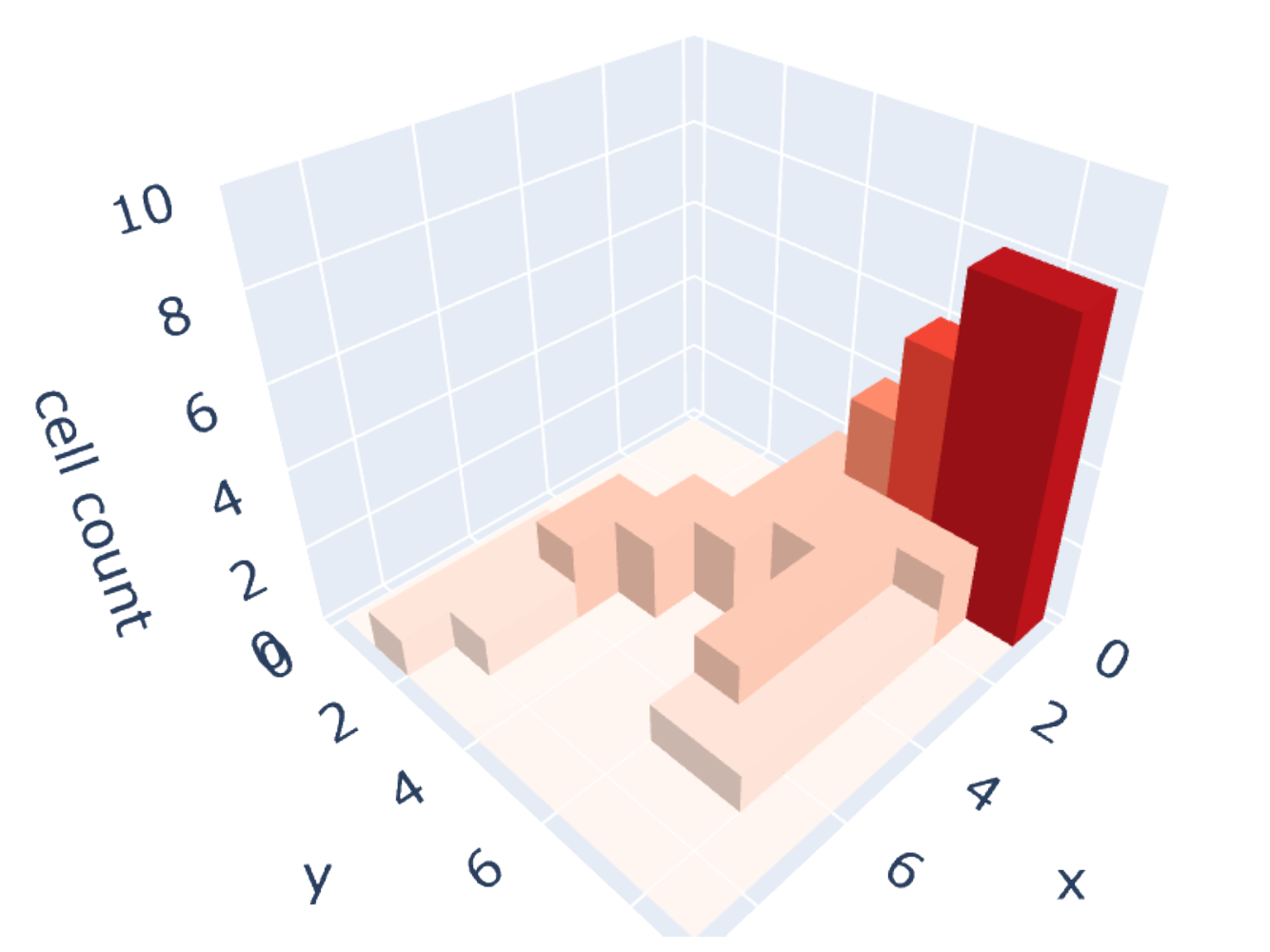}}
\subfloat[REACT $\mathcal{T}$\label{fig:ReactT-HoleyGrid}]{\includegraphics[width=0.2\linewidth]{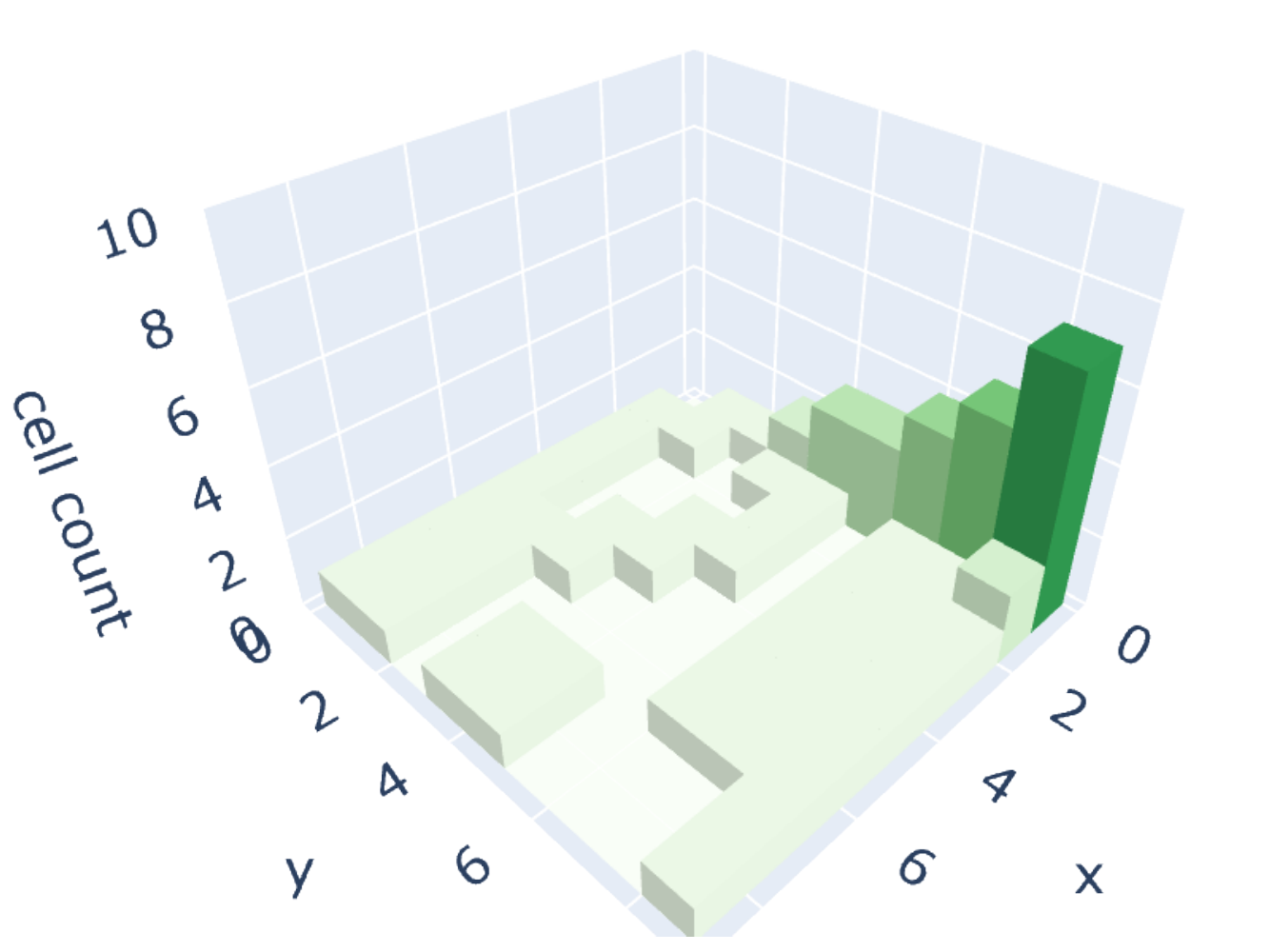}}
\caption{HoleyGrid Evaluation: {\normalfont Comparison of the Final Return~\subref{fig:HoleyGrid-Return} and Length~\subref{fig:HoleyGrid-Length} of Random~\subref{fig:RandomT-HoleyGrid} and REACT~\subref{fig:ReactT-HoleyGrid} demonstrations of a PPO policy trained for 150k steps in the HoleyGrid11~\subref{fig:HoelyGrid11}. The training performance in the unaltered environment is displayed by a solid line.
The plots indicate further edge-case demonstration being generated using REACT over random or static initial states. }} \label{fig:Eval-HoleyGrid}
\Description{HoleyGrid Evaluation}

\end{figure*}
To accurately show the influence of the \textit{local diversity} (light blue) and the \textit{certainty} (orange), we visualize their population-distance, which is combined in the minimum \textit{local distance} (yellow) to be accumulated with the \textit{global diversity} (blue) (cf. Eq.~\eqref{eq:joint_fitness}).
Interestingly, already within a population of size $10$, individual fitness decreases throughout the population, reaffirming the chosen population size.
Individuals that are evaluated with a lower global fitness (baring higher similarity to the overall population) show higher local distances, i.e., dissimilarities to the population regarding diversity of the behavior itself, which conceptually justifies considering both diversity perspectives.
In addition, all fitness components are shown to influence the whole behavior optimization, evenly increasing throughout the 40 generations. 
The considerably minor improvement in the last 10 generations indicates convergence of the optimized demonstrations.

\section{Evaluation}

\subsection{Holey Grid}
To further evaluate our approach, we use the more complex \textit{HoleyGrid} environment from \cite{Altmann_hyphi_gym} shown in Fig.~\ref{fig:HoelyGrid11}, extending the previous \textit{FlatGrid} with holes that immediately terminate an episode with a reward of $-50$. 
The holes add additional complexity to the gridworld, since the policy needs to learn to circumvent them to successfully reach the target state. 
The policy to be analyzed is trained with PPO for 150k steps in a static layout, just reaching successful behavior, with a return of $36$ and a trajectory length of $14$, as shown in Figs.~\ref{fig:HoleyGrid-Return} and \subref{fig:HoleyGrid-Length}.

The evaluation results in Fig.~\ref{fig:Eval-HoleyGrid} reveal an smaller range of returns and trajectory lengths compared to the FlatGrid results, presumably caused by the additional holes. 
In contrast to the unaltered training environment in which the policy navigates successfully, we are able to reveal unsuccessful behavior with returns slightly below $-50$. 
Again, REACT covers a slightly larger fraction of both the return and the length of trajectories compared to demonstrations from randomly generated initial states.
This is also reflected in the demonstration 3D histograms in Figs.~\ref{fig:RandomT-HoleyGrid} and \subref{fig:ReactT-HoleyGrid}.
REACT demonstrations almost cover the whole state space, where, due to the nature of the fitness, we can assume all remaining states to yield comparable behavior that would not increase the demonstration diversity. 
The unoptimized demonstrations only cover more direct solution paths, which is also reflected in the smaller interquartile range of the according returns.  
For an in-depth analysis of optimization progress and the impact of the joint fitness, please refer to the appendix. 

Overall, combining the high demonstration coverage of 10 highly diverse yet comprehensibly compact trajectories, we argue that REACT would allow a human to properly assess the inherent behavior of the trained policy.
Concretely, the analyzed policy could be described robust with a high certainty, given the above-zero interquartile range of the demonstration returns, where, depending on the intended application, further training in some problematic edge cases could be desirable.
Yet, especially compared to a single (optimal) training trajectory randomly chosen initial positions, REACT increases the interpreatbility of the policy at hand.
Note that we expect the random-sampling approach to poorly scale when introducing further disturbances (e.g., the target position or layout), where evolutionary approaches are known to efficiently cover increasing search spaces. 

\begin{figure*}\centering
\subfloat[FetchReach\label{fig:FetchReach-Env}]{\includegraphics[width=0.15\linewidth]{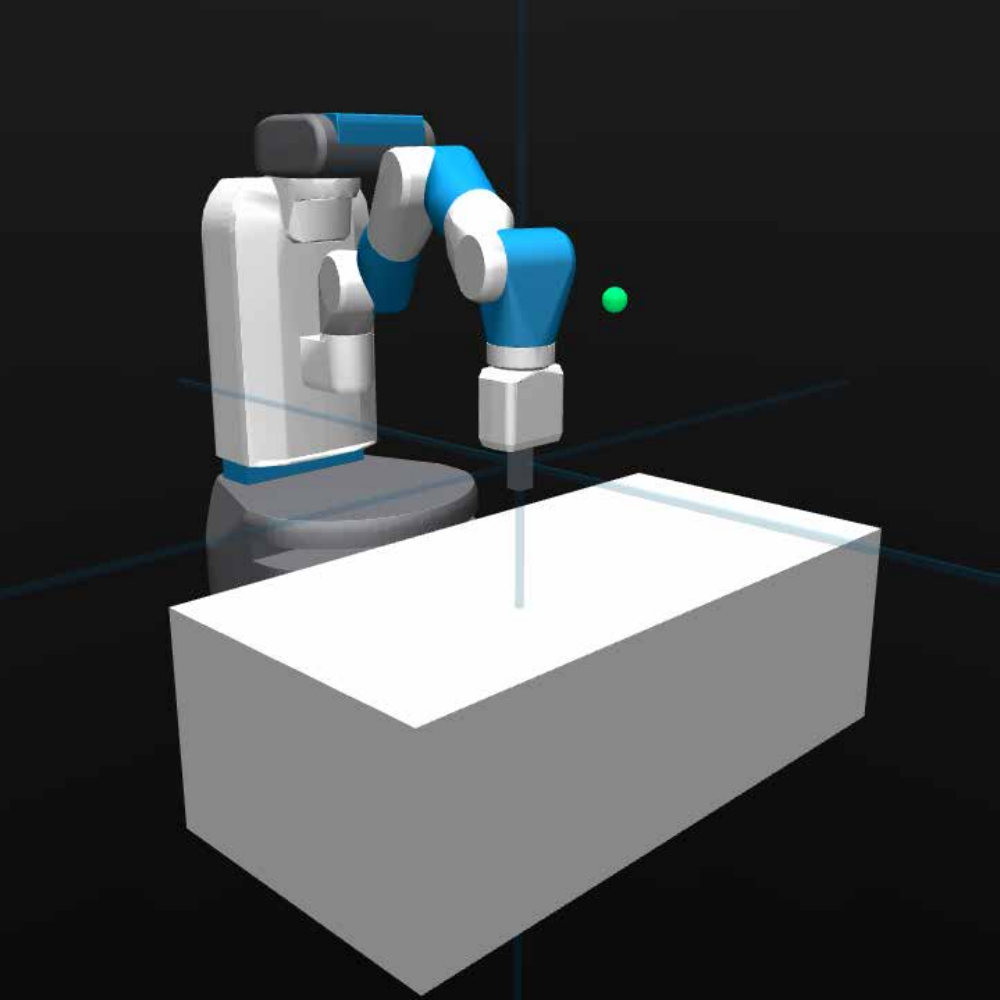}}
\subfloat[Final Return\label{fig:FetchReach-Return}]{\includegraphics[width=0.2\linewidth]{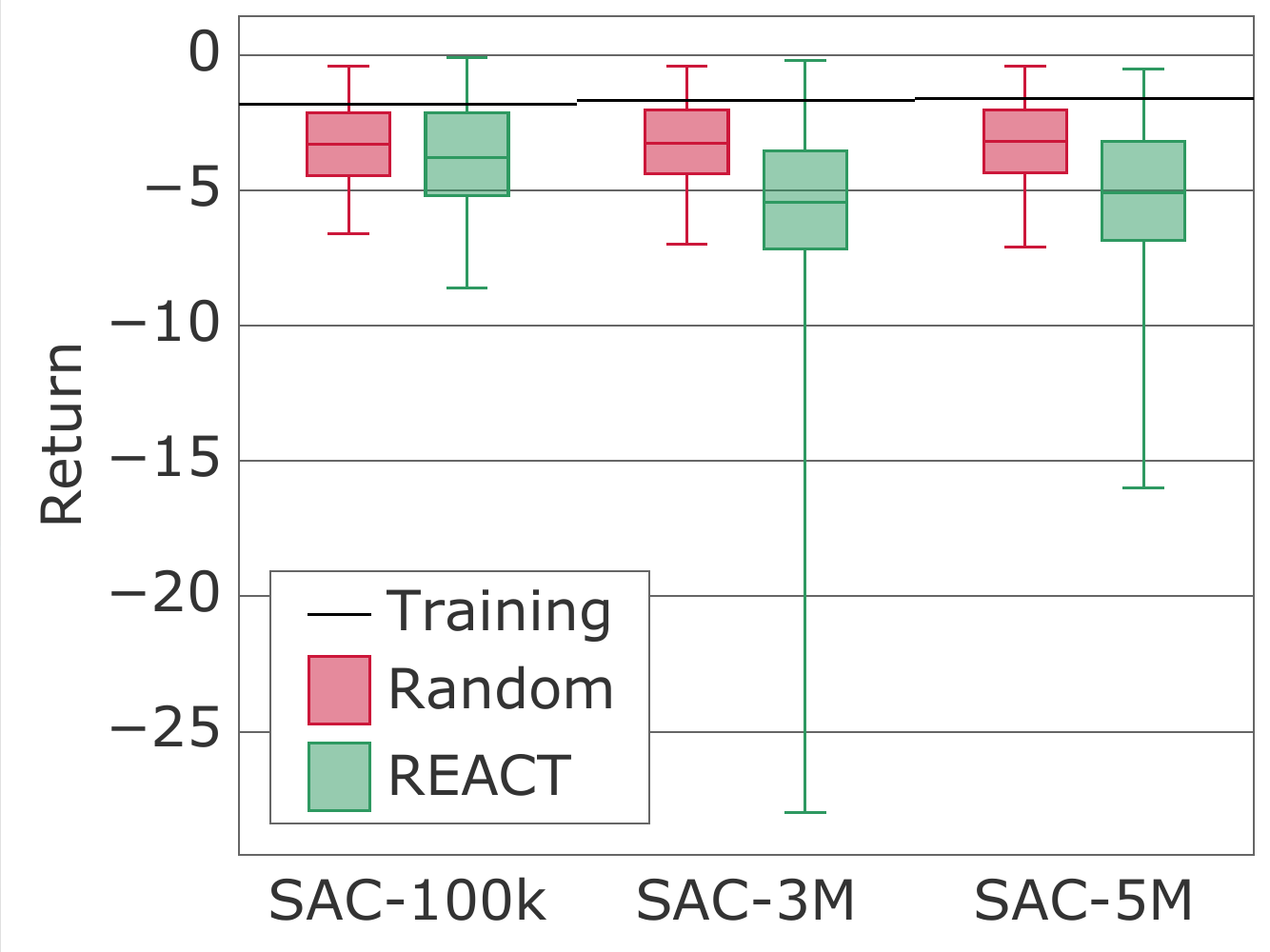}}
\subfloat[Final Length\label{fig:FetchReach-Length}]{\includegraphics[width=0.2\linewidth]{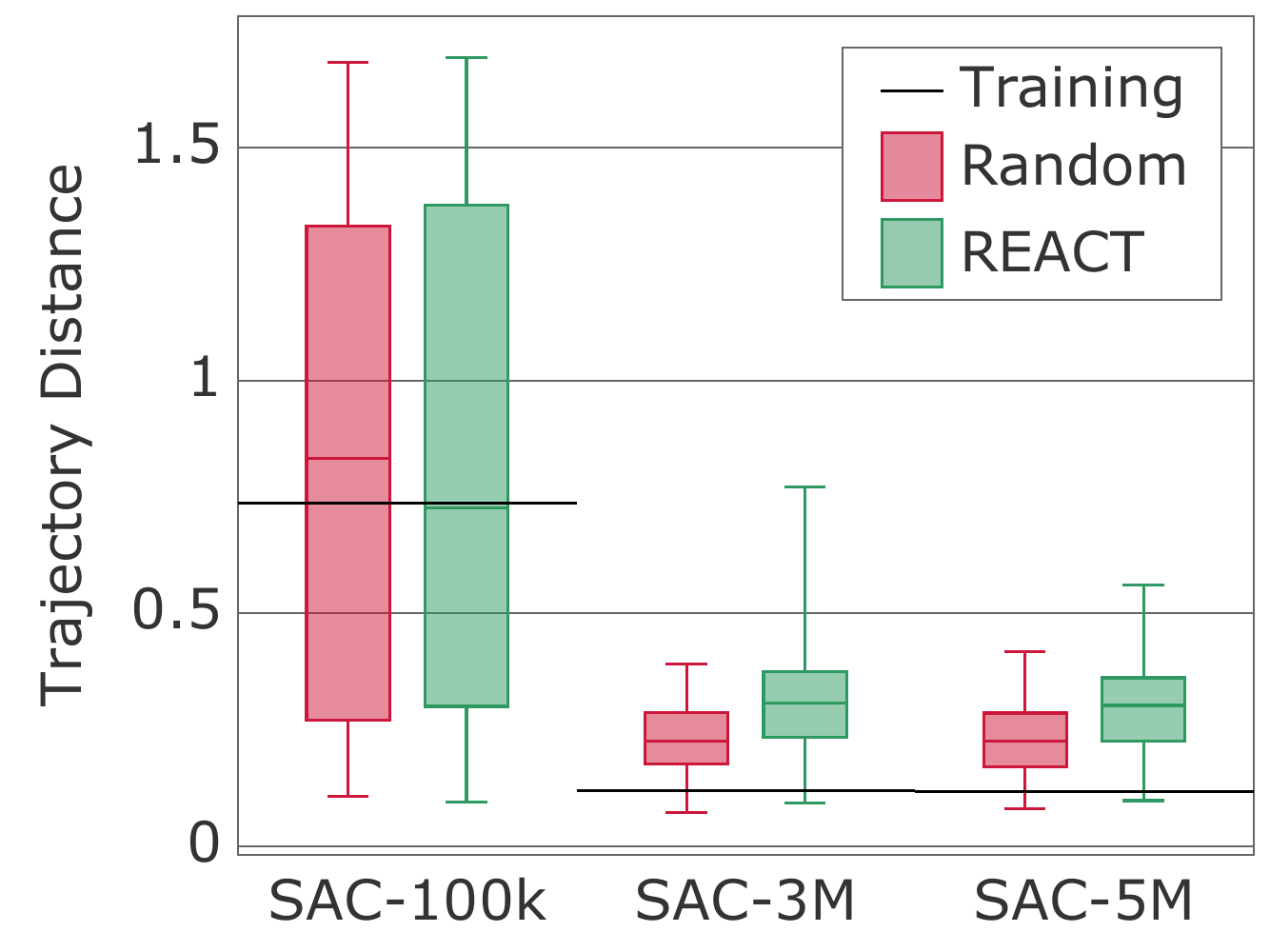}}
\subfloat[SAC-100k $\mathcal{T}$\label{fig:FetchReach-SAC100k-T}]{\includegraphics[width=0.15\linewidth]{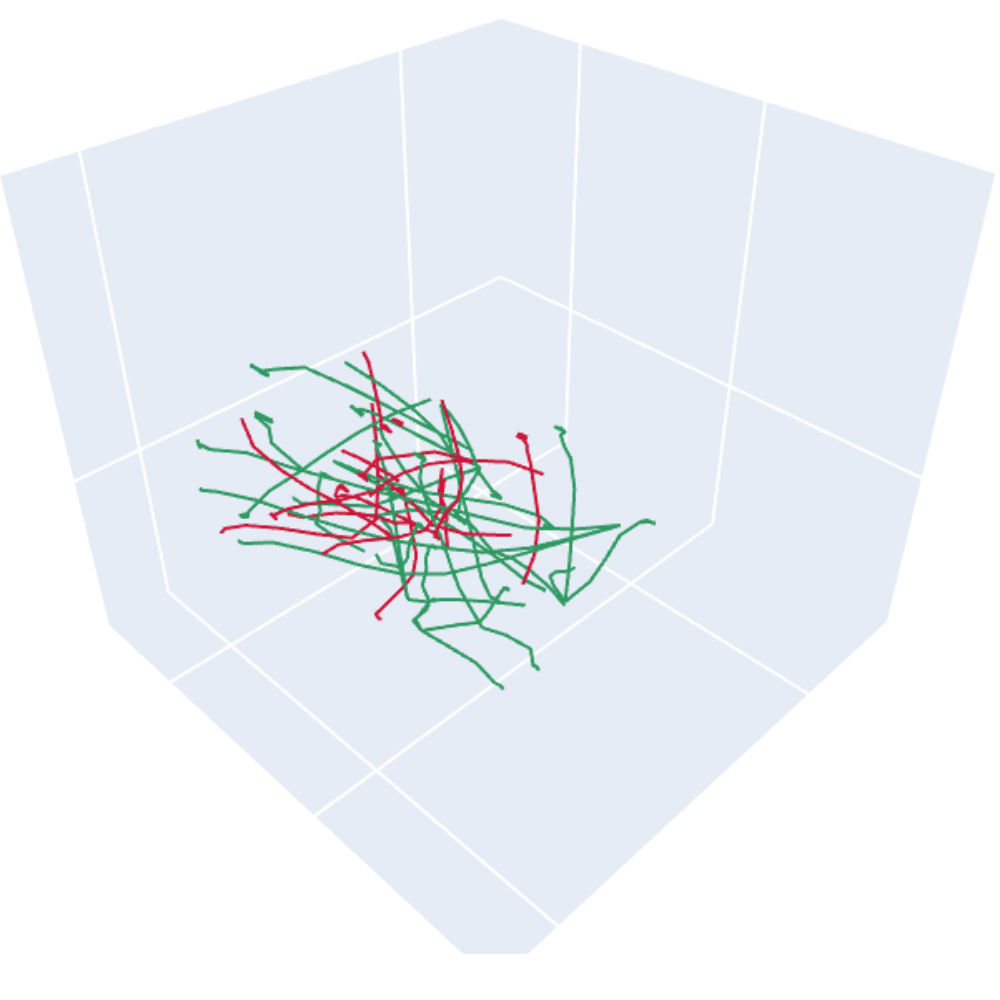}}
\subfloat[SAC-3M $\mathcal{T}$\label{fig:FetchReach-SAC3M-T}]{\includegraphics[width=0.15\linewidth]{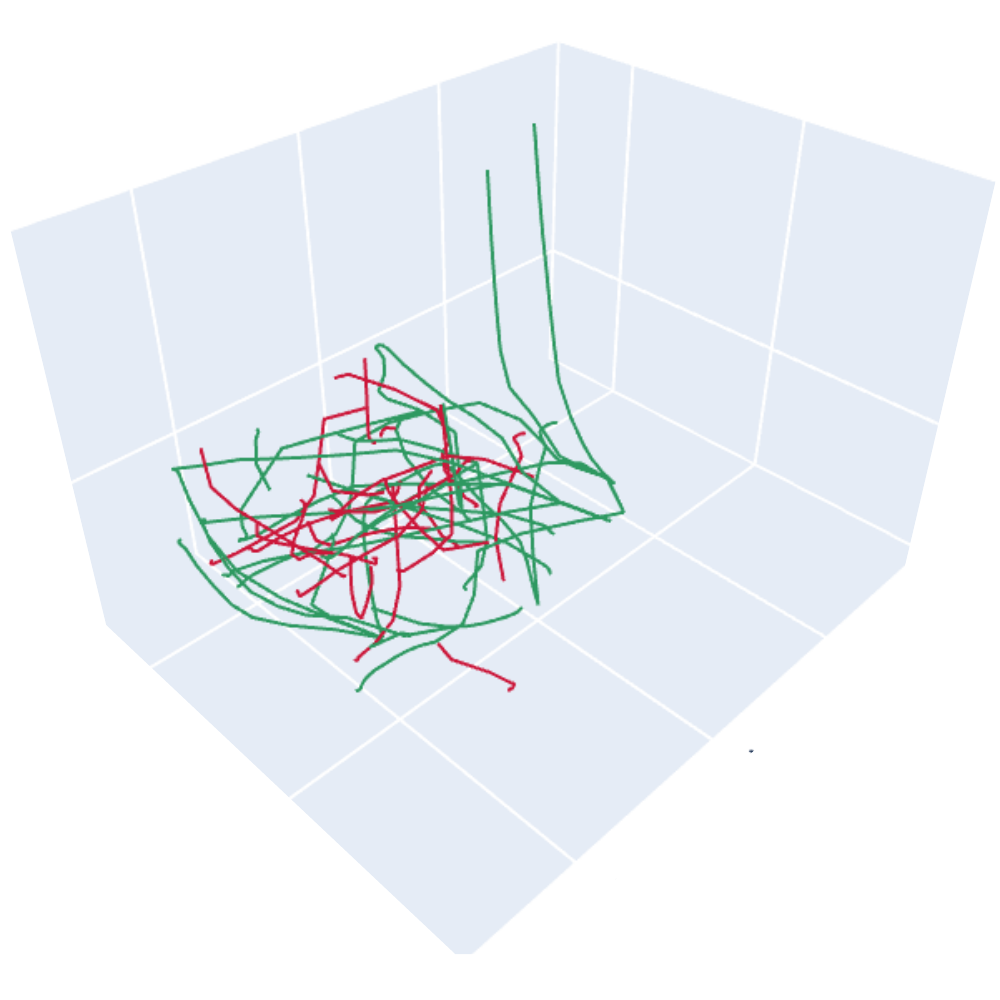}}
\subfloat[SAC-5m $\mathcal{T}$\label{fig:FetchReach-SAC5M-T}]{\includegraphics[width=0.15\linewidth]{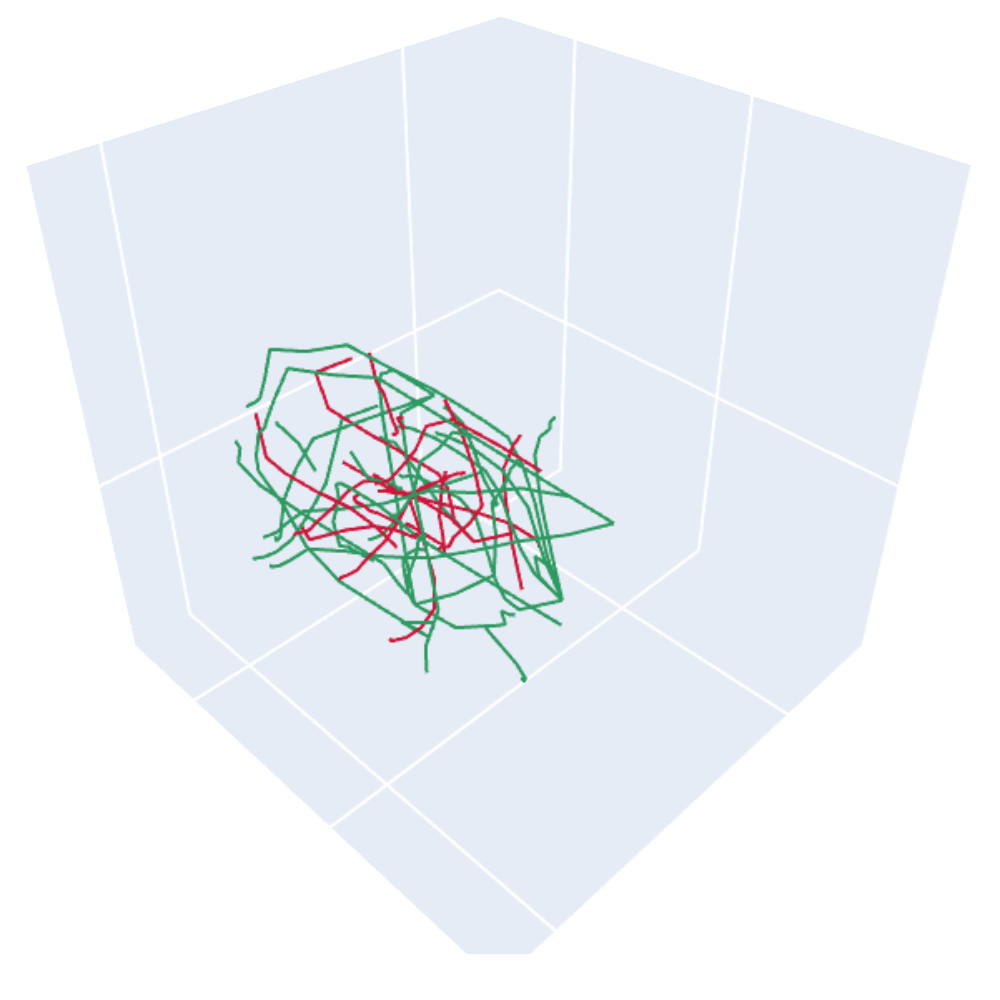}}
\caption{REACT Evaluation: {\normalfont Comparison of the Final Return~\subref{fig:FetchReach-Return} and Length~\subref{fig:FetchReach-Length} of Random (Red) and REACT (Green) demonstrations of SAC policies trained in the FetchReach~\subref{fig:FetchReach-Env} environment for 100k~\subref{fig:FetchReach-SAC100k-T}, 3M~\subref{fig:FetchReach-SAC3M-T}, and 5M~\subref{fig:FetchReach-SAC5M-T} steps.
Altogether, the plots demonstrate the applicability of REACT to discern policies from different training stages by disclosing their inherent behavior.}} \label{fig:Eval-FetchReach}
\Description{REACT Evaluation}
\end{figure*}

\subsection{FetchReach}

Finally, we demonstrate the effect of REACT in a more complex real-world application, where it could be utilized to decide between the deployment of different policies.
For this, we use \textit{FetchReach}, a continuous robotic control environment from \cite{gymnasium_robotics2023github} visualized in Fig.~\ref{fig:FetchReach-Env}.

\paragraph{\textbf{Environment}}
The agent is represented by a manipulator, the robotic arm, with six degrees of freedom, and its end effector, a gripper. 
The task is to control the robotic arm by applying a three-dimensional force vector to move the gripper to reach the target state (green point). 
In contrast to the previous gridworld environments, both action- and observation-space are real-valued. 
Furthermore, the task is open-ended such that episodes continue for 50 steps regardless of successfully reaching the target. 
Therefore, we use a sparse reward function, where the agent is penalized $-1$ for every step in which it is not close to the target, i.e., where the Euclidean distance between the effector and the target is greater than $0.05$.
During training, the effector's position is always initialized at the center, while the target is randomly positioned within a $0.3$-sized cube around the center to improve generalization of the learned behavior \cite{Gymnasium-Robotics_Contributors_Gymnasium-Robotics_A_a_2022}. 

\paragraph{\textbf{REACT Parameters}}

Given the increased environmental complexity, we adapted the parameterization of REACT according to preliminary studies. 
Most importantly, to remove all random factors from the generation of demonstrations, we include the target position in addition to the agent (gripper) position in the initial state to be optimized.
This results in a 6-dimensional state encoding, which we chose to encode with a bit-length of 9 to reduce the intervals between possible states to less than $0.001$.
Furthermore, we increased the population size to $30$ and used $1000$ generations.
Lastly, instead of analyzing a single moderately trained policy, we compare polices from three stages of training.

\paragraph{\textbf{Training}}
Having proven beneficial in various continuous control tasks, we train the policies to be compared using \textit{SAC} \cite{SAC}, implemented with default parameterization \cite{stable-baselines3}.
To demonstrate the comparative evaluation capabilities, we trained policies for 100k, 3M, and 5M steps, we refer to as \textit{SAC-100k}, \textit{SAC-3M}, and \textit{SAC-5M} respectively for the following. 
Therefore, we are able to compare policies from three training-stages, ranging from early convergence to possibly over-trained, thus overfitting the training task.

\paragraph{\textbf{Results}}
The overall evaluation results are shown in Fig.~\ref{fig:Eval-FetchReach}. 
Looking at the performance of all models in the unaltered training environment shows both increasing returns of around $-1.8$, $-1.7$, and $-1.6$ and decreasing trajectory distances of around $0.73$, $0.12$, and $0.11$ for SAC-100k, SAC-3M, and SAC-5M respectively.
With a maximum target-distance of $0.3$, based only on these results, the primal SAC-100k could be disregarded due to the significantly more extensive movement, even though reaching competitive rewards.  
However, REACT reveals further important insights to base model interpretations and subsequent decisions on. 
Regarding the final trajectory length, REACT shows diverse demonstrations to be evenly distributed around the single training experience for SAC-100k, with both the length and the variance of the length decreasing upon further training. 
Compared to demonstrations based on random initial states, REACT again shows to slightly increase the overall diversity and even distribution of demonstrations. 
More interesting results, however, are shown for the final return, where random configurations, similar to the training configuration, do not reveal any insightful differences between the models. 
REACT, on the other hand, reveals the overall return variance increasing with further training, with the median of returns even decreasing. 
It is important to note that the return is not included in the fitness to optimize the demonstrations. 
Thus, these observations emerge from diverse behavior generated by the policies.
Given the increasing returns observed for the training configuration, this could most likely be explained as overfitting behavior.
To give an intuition of the scope and nature of the resulting demonstrations, we finally consider the path of all Random (red) and REACT (green) trajectories shown in Figs.~\ref{fig:FetchReach-SAC100k-T}-\subref{fig:FetchReach-SAC5M-T}.
Due to the continuous nature of the environment, combined with the increased number of individuals, we did not plot the resulting demonstrations as cumulative distributions (mainly due to the fact that averaging would diminish any diversity within the populations). 
Although this kind of visualization does not allow for the precise analysis of each resulting trajectory, it perfectly conveys the overall nature of the generated demonstrations.
Again, REACT covers a comparably larger fraction of the state space more evenly, and even detected a policy insufficiency of SAC-3M causing the demonstration of an outlier. 
In summary, the shortest-trained policy reaches targets the fastest, showing the lowest penalties and thus the highest returns, but, with the lowest precision, thus, highest movement and trajectory length.
Longer-trained policies, on the other hand, show higher penalties, reaching the target slower, yet more precise, as indicated by the overall lower trajectory length. 
The assessment of those characteristics heavily depends on the intended application; however, REACT has shown to reveal those important characteristics of the inherently learned behavior.

\section{Conclusion}

To enhance the interpretability of RL, we introduced \textit{Revealing Evolutionary Action Consequence Trajectories} (REACT). 
REACT adds disturbances to the environment by altering the initial state, causing the policy to generate edge-case demonstrations.
To assess trajectories for demonstrating a given policy, we formalized a joint fitness combining the local diversity and certainty of the trajectory itself with the global diversity of a population of demonstrations. 
To optimize a pool of demonstrations, we apply an evolutionary process to the population of individuals, encoded as the initial state, evaluated by the joint fitness. 
To evaluate REACT, we analyzed various policies trained in flat and holey gridworlds as well as a continuous robotic control task at different training states. 
Comparisons to the unaltered training environment and randomly generated initial states showed that REACT reveals a set of more diverse and more evenly distributed demonstrations to serve as a varietal basis to assess the learned (inherent) behavior. 
We evaluated the results based on their final return, which is intentionally precluded from the fitness calculation.
However, we refrained from analyzing the demonstrations' utility for a human assessment of the policy: we consider this inherently task-specific and therefore out-of-scope, as we envision REACT to be a task-agnostic approach. 
Furthermore, we only introduced disturbances of the initial agent and target positions. 
Thus, future work should examine extending REACT to further variations of the environment, such as the overall layout or the task itself. 
Also, the resulting pool of demonstrations could be used either to further improve the policy regarding revealed vulnerabilities, or to infer a global causality model to further foster the policy's interpretability. 
Overall, we believe that REACT represents a universal policy-centric starting point for improving the overall interpretability of the currently mostly opaque RL models. 

\begin{acks}
This work was partially funded by the Bavarian Ministry for Economic Affairs, Regional Development and Energy as part of a project to support the thematic development of the Institute for Cognitive Systems.
\end{acks}

\balance

\clearpage\appendix 
\section{State Encoding}\label{sec:state_encoding}

In the following, we elaborate our encoding design of the initial state $s_0$, defining the individual's genotype.
The initial state is defined by the agent's initial position. 
In most environments, the initial state is either fixed or randomly chosen (by sampling from $\mu$), as is the goal state. 
Fixed and randomly selected initial states and goals have a high impact on the behavior, a trained agent is going to display. 
For this reason, initial states and goals should be taken into account when deciding the configuration of the environment.
This has the benefit of controlling random behavior of the agent by reducing random factors in the environment. 
Additionally, it creates the possibility of an unlikely initial state for the agent to start in. 
This gives us the chance to see the agent act in states, that would otherwise not be encountered in a normal episode.

\begin{figure}[ht]
    \centering
    \includegraphics[width=0.75\linewidth]{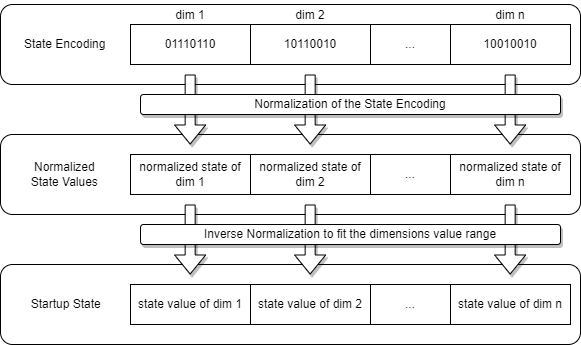}
    \caption{Mapping of State Encoding to Startup State}
    \label{fig:encoding}
    \Description{Mapping of State Encoding to Startup State}
    
\end{figure}

We used a binary encoding to represent the altered initial state, since it represents a simple way of encoding information and can easily be used for recombination and mutation operators.
However, the challenge we face with a binary encoding is that with $m$ bits, exactly $2^m$ possible states can be represented. 
Considering our example gridworld with a 9x9 grid, we have 81 states representing possible intial positions.
A binary encoding of 6 bits could only represent 64 states while 7 bits can represent 128 states. 
We solved this challange by appling inverse normalization.
Every dimension $d$ of the startup state space is represented by a binary encoding $e$ of length $m$.
As shown in \ref{fig:encoding}, we first map each part of the binary encoding to an integer, followed by normalizing these integers. 
The normalized values are then mapped to possible states of their respective dimensions.
The state values build the final initial state for the environment. 

The benefit of this mapping from binary encoding to intial state value is that we can represent discrete state spaces with different sizes, as well as continuous state spaces with a specified precision. 
For discrete and continuous state spaces, we have a slightly different calculation of the normalization of the state encoding and the inverse normalization to the actual state value.

\paragraph{Discrete State Space}

Cutting the state encoding in $\vert d \vert$ equally sized encodings $e_d$, we can compute the corresponding value as follows:
The binary encoding $e_d$ with length $m$ is translated to its corresponding integer value. This integer value is then normalized by dividing it by the number of possible integer values for the encoding length.
$$normalizedState = \frac{int(e_d)}{2^m} $$
The $normalizedState$ is a real number in $[0,1[$. In the next step it will be mapped to the dimensions value range.
To map the normalizedState to an integer, that lies in the dimensions value range, we require the maximal and the minimal possible values of the dimensions state space. We then reverse the normalization using the new value range and receive a real number, that is then rounded off to the next lower integer value.
$$state = \lfloor normalizedState * (maxValue +1 - minValue) + minValue \rfloor$$
In a discrete state space, we have to choose an encoding length which can represent more states, than are actually present in the state space. This ensures, that each state can be represented by at least one encoding. 
Since we cannot map the states of the encodings uniformly to the states of the state space, we accept the compromise that there are slightly different probabilities of occurring for the different states. By choosing a longer encoding, the difference in the probabilities of occurrence between the states is reduced.
We can compute the difference in probabilities of occurrences depending on the chosen encoding length and the value range of the mapped dimension.
$$intervalRange = \frac{maxValue + 1 - minValue}{2^m}$$
$$lowerProbability = \frac{\lfloor \frac{1}{intervalRange} \rfloor}{2^m} $$
$$higherProbability = \frac{\lceil \frac{1}{intervalRange} \rceil}{2^m} $$

For our gridworld environment, this means that for a grid of 9x9 cells, the initial position of the agent would be encoded by two encodings of at least length 4. Each encoding respresents one dimension of the gridworld, the row and the column, which each have 9 possible states with $maxValue = 8$ and $minValue = 0$.

\paragraph{Continuous State Space}

For a continuous state space, we do not require the state values to be integers. For this reason, we have a slightly different calculation.
Here, we choose the encoding length based on the precision of the state values that we want to receive.
The state encoding is normalized as follows:
$$normalizedState = \frac{int(e_d)}{2^m - 1}$$ 
The difference to the calculation in the discrete space is a range of $[0,1]$ for the $normalizedState$.
In the inverse normalization, we do not round the value off, which is why we do not need to add $1$ to the $maxValue$.
$$state = normalizedState * (maxValue - minValue) + minValue$$
For continuous state spaces, a longer encoding equals a higher precision. 
We can calculate the difference between two encodings values as follows:
$$stateValueDistance = \frac{maxValue - minValue}{2^m - 1}$$
\clearpage

\section{Additional Experimental Results}

\subsection{Encoding Length}

We already approached the topic of choosing the right encoding length in \autoref{sec:state_encoding}.
The encoding of each dimension has to have a sufficient length such that every state in the state space can be represented by it.
However, in an environment with a discrete state space, some startup states could be represented by more encodings than other startup states. 
With our approach, this leads to a higher probability of occurring. 
The difference in probabilities depends on the length of the state encoding. 
Considering a grid size of $11\times11$, the possible startup state are only represented by a grid size of 9x9. 
We did not consider cells on the surrounding wall as possible startup states. Each of the two dimensions has 9 possible values, that need to be encoded by a binary encoding with at least 4 bits.
We concluded a simple experiment, to be able to better recommend an encoding length.
We compared five different encoding lengths per dimension: $n \in [4, 5, 6, 7, 8]$.
For each encoding length, we created 81000 random bit encodings of length $2*n$. We transformed them to states in the environment using the method described in \autoref{sec:state_encoding} and plotted them in a heatmap to be able to show the difference in the occurrence of states.
\begin{figure}[ht]\centering\vspace{-0.1cm}
\includegraphics[width=.9\linewidth]{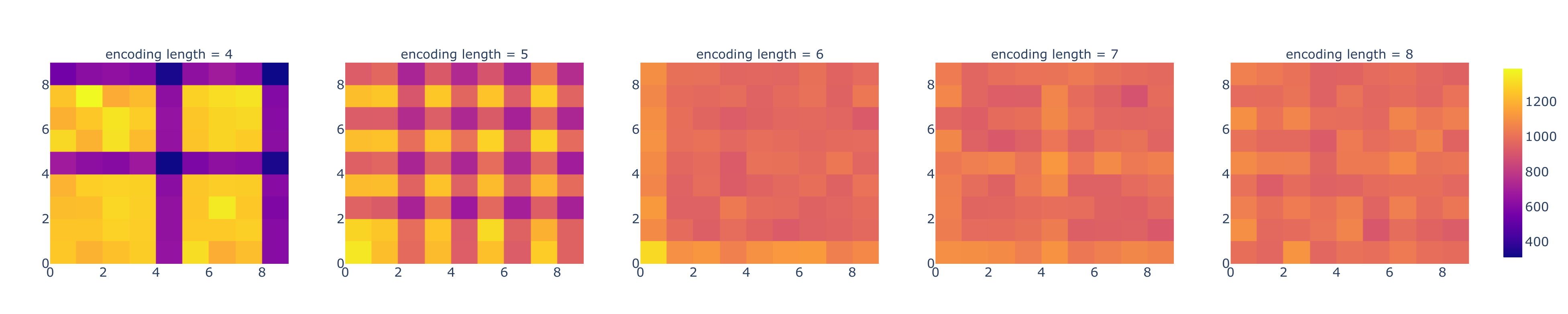}
\vspace{-0.3cm}\caption{Comparing 81000 random 11x11 state encodings}\label{fig:enc_length_comparison}
\Description{State encoding comparison}
\end{figure}
Looking at Figure \ref{fig:enc_length_comparison}, we can see a significant difference in the length of encodings.
Especially for an encoding length of 4 or 5, the difference in occurrence is clearly visible. The darker cells are less probable to occur then the lighter cells. This supports the computation of the difference in probability described in \autoref{sec:state_encoding}. For an encoding length of 4, some states are 2 times as probable as others. For an encoding length of 5, this is reduced to 1.33 times. Looking at the heatmaps and the calculation of the difference in probability, we recommend to use a state encoding length, where a state is less than 1.25 more likely than another state. In FlatGrid11, this is the case for an encoding length of 6. There only exist a few states that are 1.14 more likely than other states. This difference in probability can be regarded as insignificant for our algorithm.

\subsection{Population Size}\label{sec:population_size}
The population size determines how many demonstrations are shown to the user. 
On one side it is important to show enough different behaviors for the user to get a good impression on how the agent behaves. 
Also, we could execute the genetic algorithm using a greater population size, but only display the 10 best individuals. 
However, this has several drawbacks.
First, a greater population size makes it easier for individuals to survive in the population. A larger population means, there will also be more individuals with a low fitness value. This also applies to individuals that are not diverse.
Second, a greater population size makes it more difficult for individuals, to be considered diverse in comparison to other individuals. In most cases the following applies: the more individuals to compare to, the less diverse the individual can be. This makes it more difficult for a new individual to have a high enough fitness value to consider it one of the best individuals in the population.

\begin{figure}[ht]
    \centering
    \subfloat[all individuals\label{subfig:popsize_iter_a}]{\includegraphics[width=0.4\linewidth]{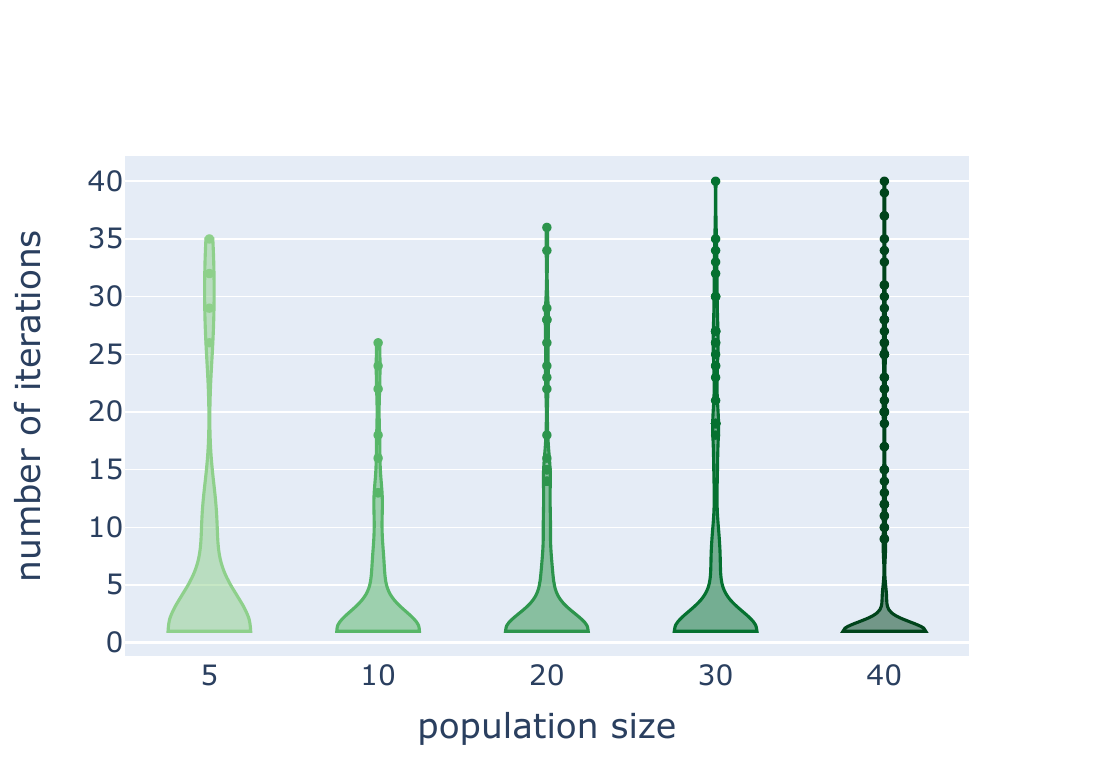}}
    \subfloat[best 10 individuals\label{subfig:popsize_iter_b}]{\includegraphics[width=0.4\linewidth]{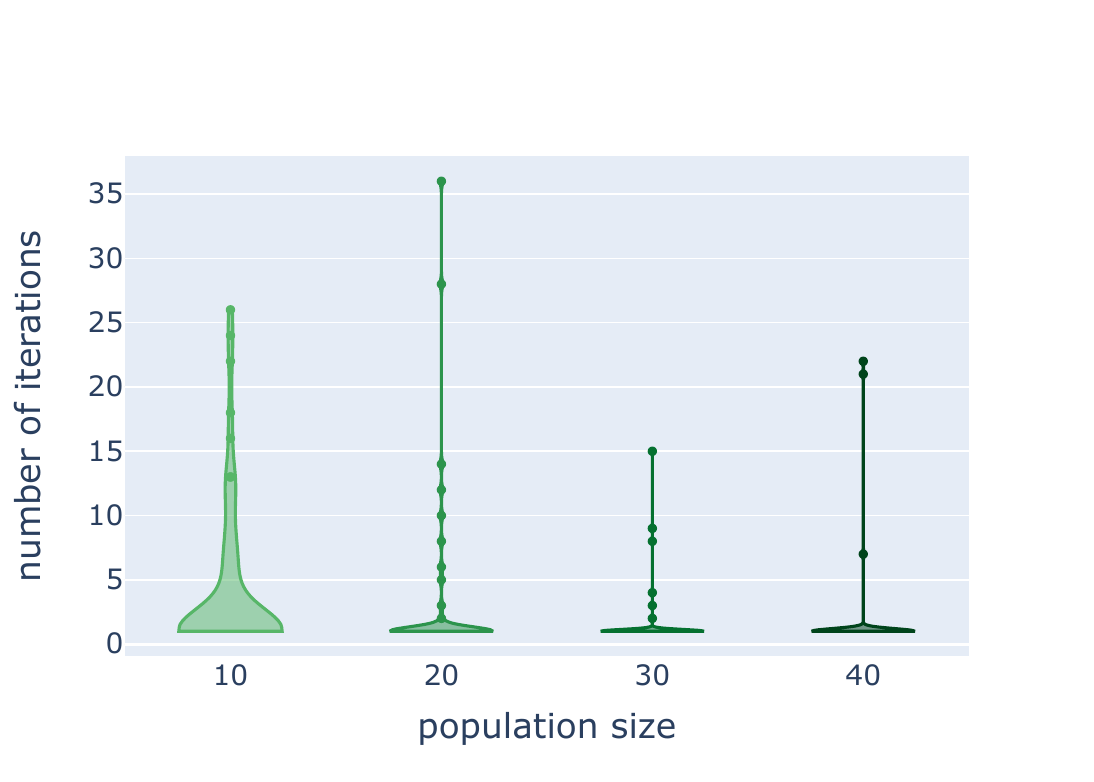}}
    \caption{New individuals joining the entire population and only the best 10 individuals of the population}
    \label{fig:popsize_iter}
    \Description{Population size evaluation}
    
\end{figure}

This reasoning is supported by Figure \ref{fig:popsize_iter}, where we can see all newly added individuals that survived at least one generation. In each one of the plots, we compared the development across different sizes of the population. The displayed data contains individuals from 4 runs of the genetic algorithm. In Figure \ref{subfig:popsize_iter_a}, we look at all individuals of the respective population size. We can determine, that with a larger population we also have more individuals survive at least one generation at a later point in the algorithm. This shows, that with a larger population new individuals have higher chances of surviving to the next generation. A larger population therefore reduces the selection pressure and leads to a slower convergence of the algorithm. 
Especially in an environment with few startup states, it is more difficult to find new individuals that are considered more diverse than already evaluated individuals. Since the diversity is based on the comparison to other individuals, the initially created individuals have an advantage over newly created individuals in later iterations. Older individuals are already considered highly diverse, while newer individuals have to be compared to them. The newer individuals have more and better competition than the initially generated individuals.
Since the user is not interested in viewing all episodes of a large population, we reduced the population of the last generation to only show the best 10 individuals. The resulting plot can be seen in Figure \ref{subfig:popsize_iter_b}. 
While looking at all individuals of the population shows more new individuals joining the population, the larger the population is, we can see a clear difference to that when only looking at the best 5 individuals. The last iteration where a new individual is ranked as one of the best 5 individuals of the population is sinking for an increasing population size. Only looking at the best individuals, we now see that with a larger population the algorithm converges significantly faster.
We should not choose a too small population size. Important information could get lost. When viewing the episodes selected by the genetic algorithm, obviously we want to see the most diverse individuals, but we also want to see a few individuals that do not have an outstanding behavior. It is important to choose a population size that does not only show the most drastically diverse behavior, but instead shows as much of the behavior as possible without showing redundant behavior.
The goal is, to see only those individuals, that are strictly necessary to present diverse behavior of the agent.
Therefore, we compared the return and trajectory length distribution of all individuals in the population, with the reward and trajectory length distribution of only the best 10 individuals. 

\subsection{Crossover and Mutation Probability}
For every problem, that should be solved by a genetic algorithm, there are two hyperparameters that need to be tuned: the crossover probability and the mutation probability. A higher crossover probability usually leads to more exploitation of individuals with a high fitness value, while a high mutation probability explores the search space to avoid getting stuck in a local optimum. By applying mutation, new diverse individuals are evaluated, that could lead to new insights.
We explored several different settings for the two probabilities: a high crossover probability of 0.9 with a low mutation probability of 0.25, a slightly lower crossover probability of 0.75 with a higher mutation probability of 0.5, and both of these settings with switched probabilities. Additionally, we added the setting with a crossover probability of 0.9 and a mutation probability of 0.4, which we used in the previous experiments.
With this experiment, we mean to investigate, if the search space requires more exploitation of the already evaluated individuals or more exploration of the search space.

\begin{figure}[ht]
    \centering
    \subfloat[distribution of the returns in the last generation\label{subfig:prob_reward}]{\includegraphics[width=0.5\linewidth]{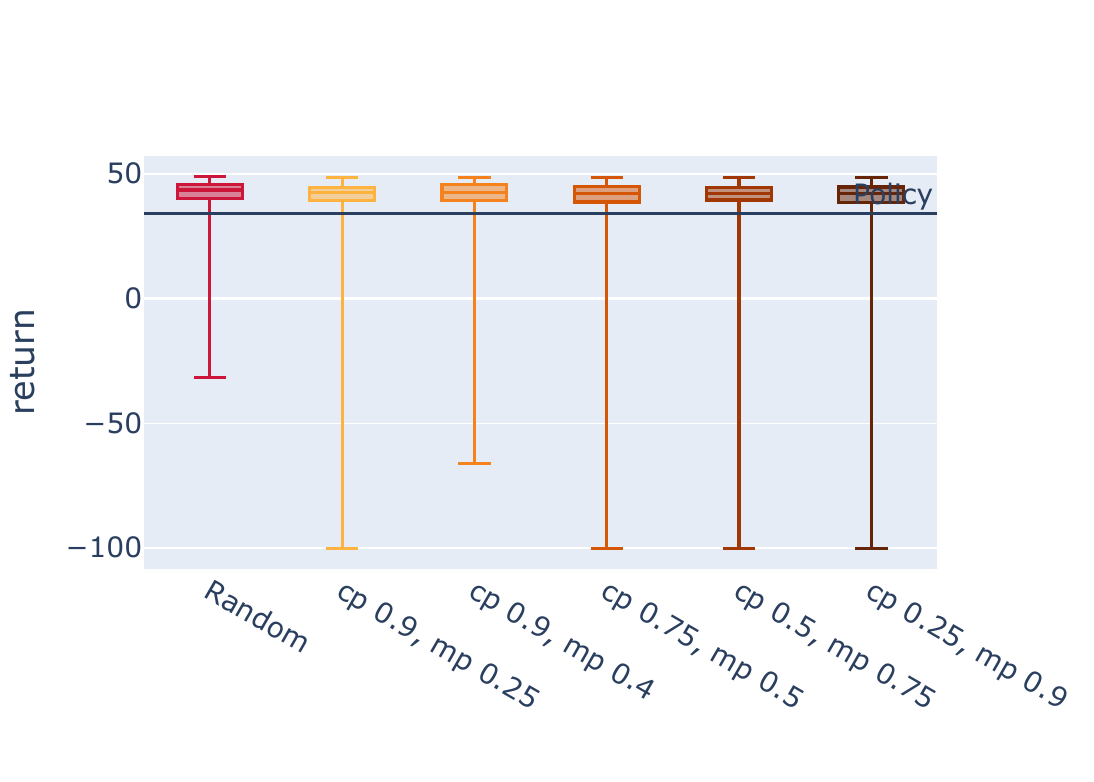}}
    \subfloat[distribution of the trajectory lengths in the last generation\label{subfig:prob_traj_len}]{\includegraphics[width=0.5\linewidth]{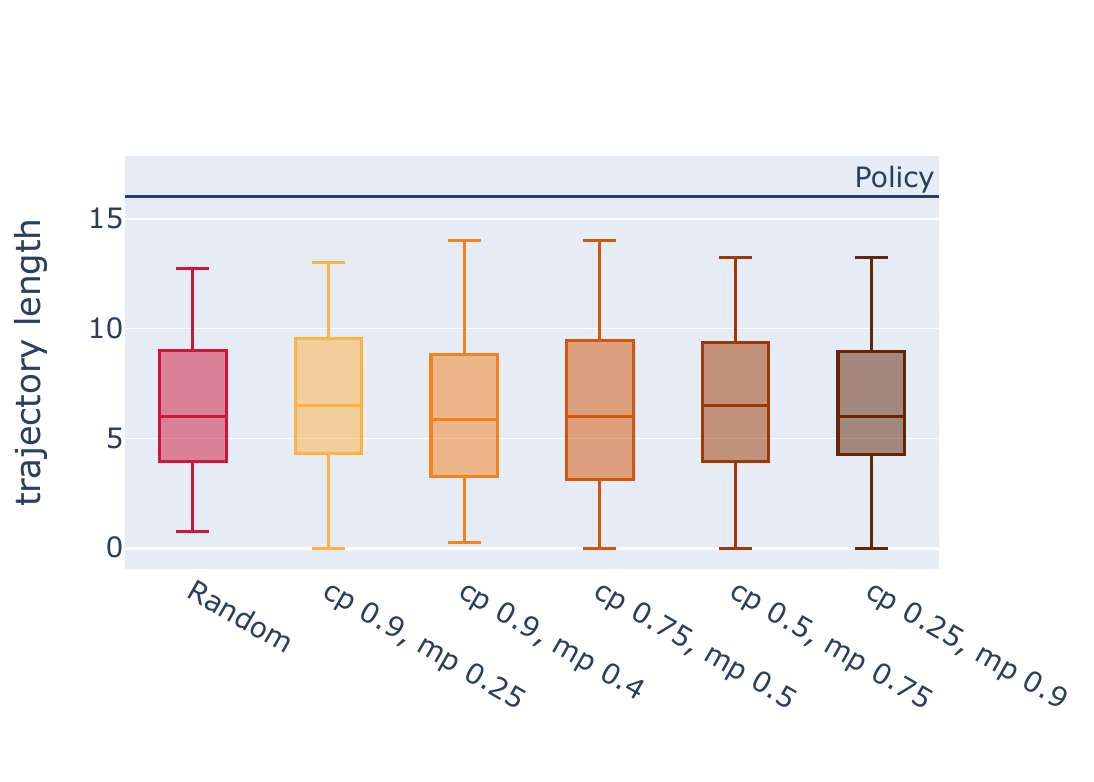}}
    \caption{Diversity of the individuals of the last generation compared by different operator probabilities}
    \label{fig:prob_results}
    \Description{Diversity evaluation}
\end{figure}

The results in Figure \ref{fig:prob_results} show, that a low crossover probability paired with a high mutation probability does not lead to significantly better results in terms of the distribution of trajectory lengths than random search. However, looking at the distribution of returns, we can state, that our algorithm shows more diverse behavior than random search. Random search does not always manage to find the startup states that lead to a very interesting behavior, namely the agent's standstill. 


\subsection{Fitness Impact Analysis}

\begin{figure}[h!]\centering
\vspace{-0.1cm}\includegraphics[width=\linewidth]{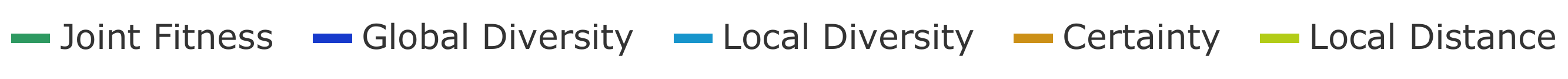}\\
\subfloat[Population Analysis\label{fig:HoleyGrid-Population}]{\includegraphics[width=0.5\linewidth]{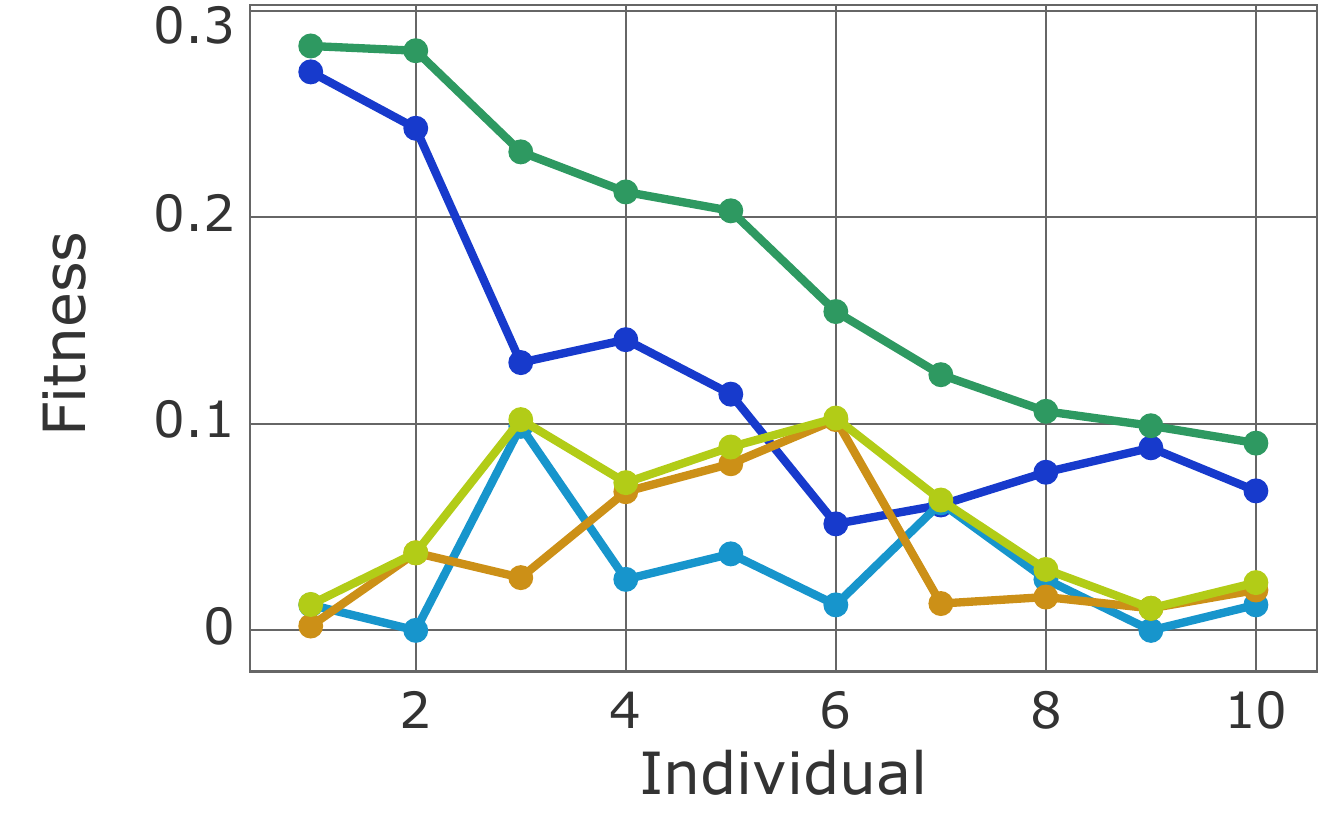}}
\subfloat[Generation Analysis\label{fig:HoleyGrid-Generations}]{\includegraphics[width=0.5\linewidth]{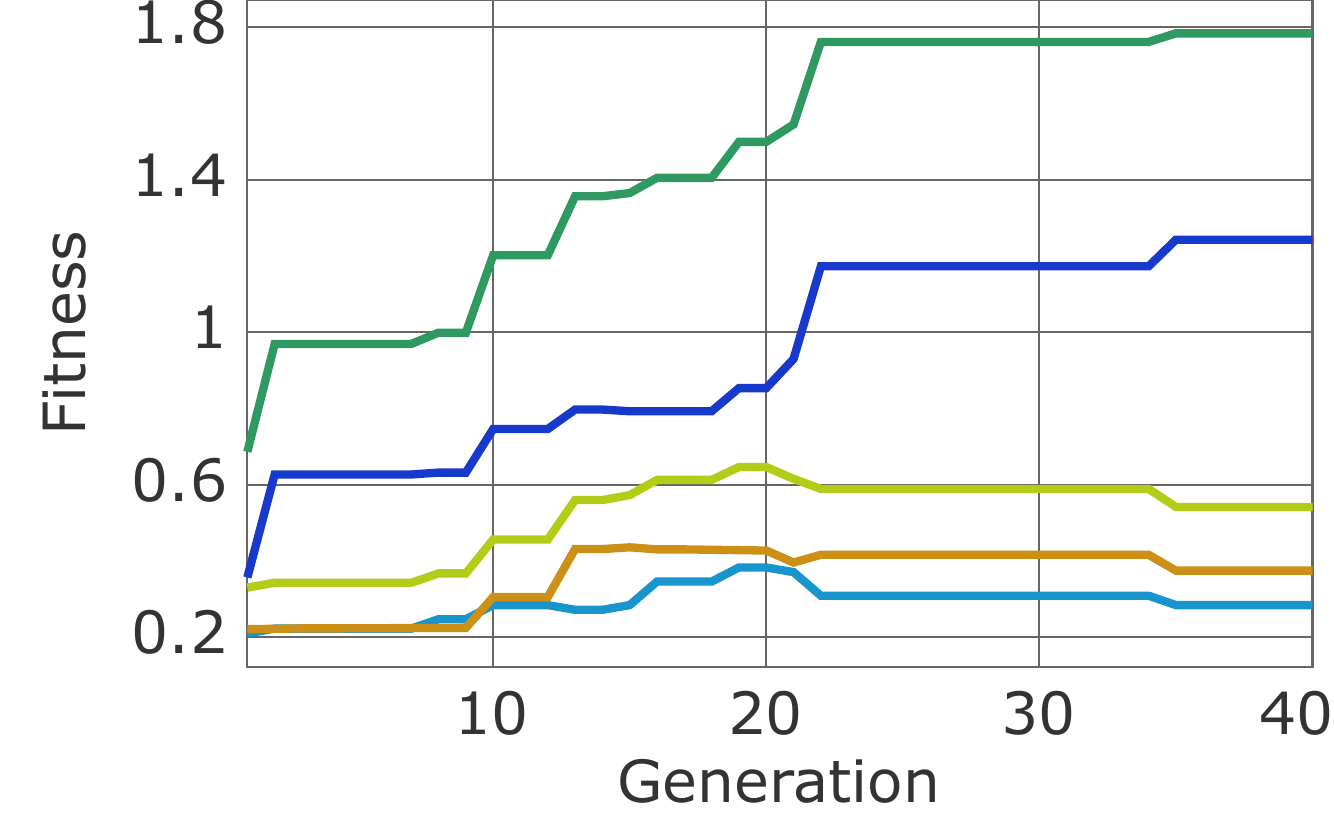}}
\vspace{-0.1cm}\caption{HoleyGrid \textit{JointFitness} Analysis\label{fig:HoleyGrid-Fitness}}
\Description{HoleyGrid \textit{JointFitness} Analysis}
\end{figure}

\begin{figure}[b]\centering
\vspace{-0.1cm}\includegraphics[width=\linewidth]{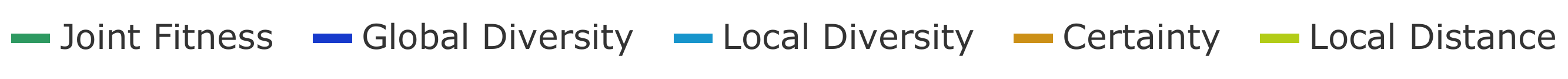}\\
\subfloat[Population Analysis\label{fig:FetchReach-Population-SAC100k}]{\includegraphics[width=0.5\linewidth]{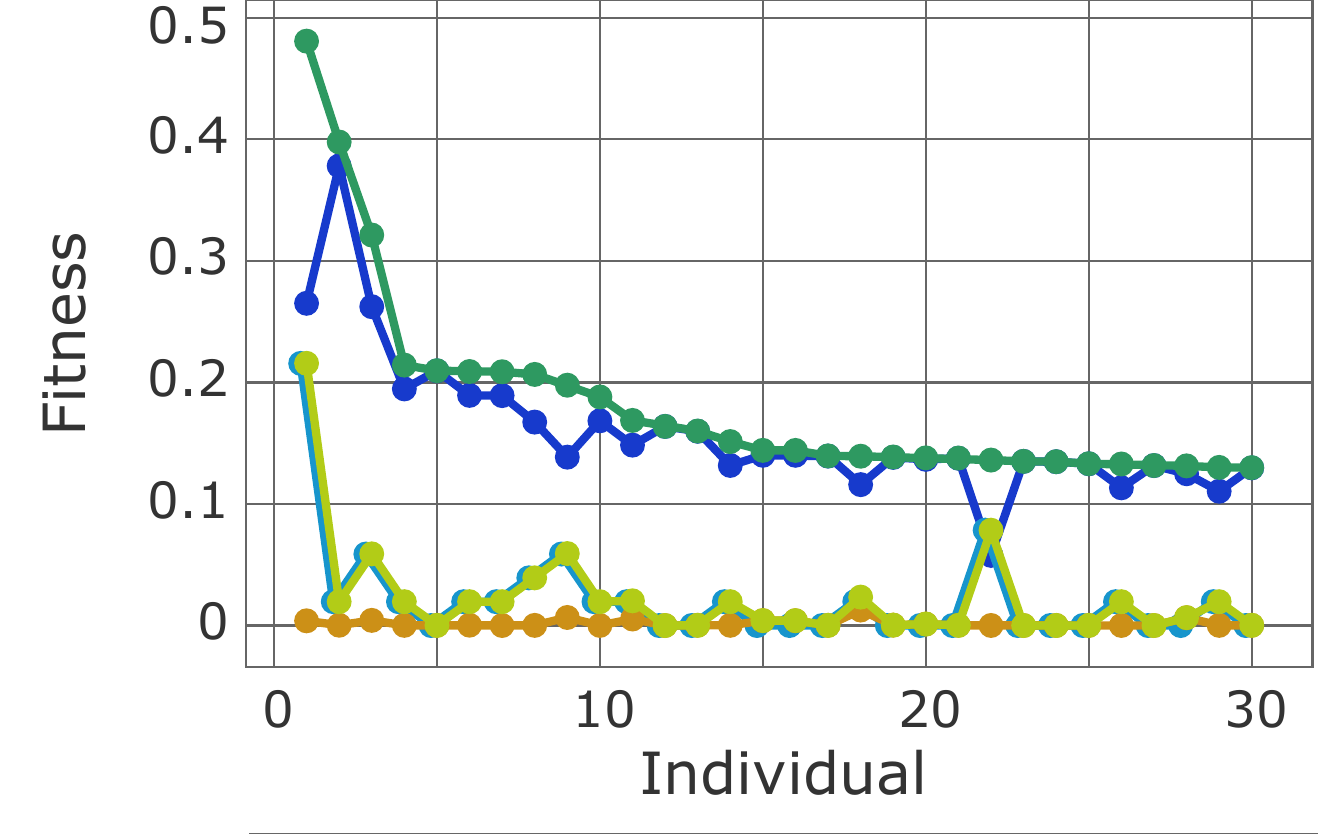}}
\subfloat[Generation Analysis\label{fig:FetchReach-Generations-SAC100k}]{\includegraphics[width=0.5\linewidth]{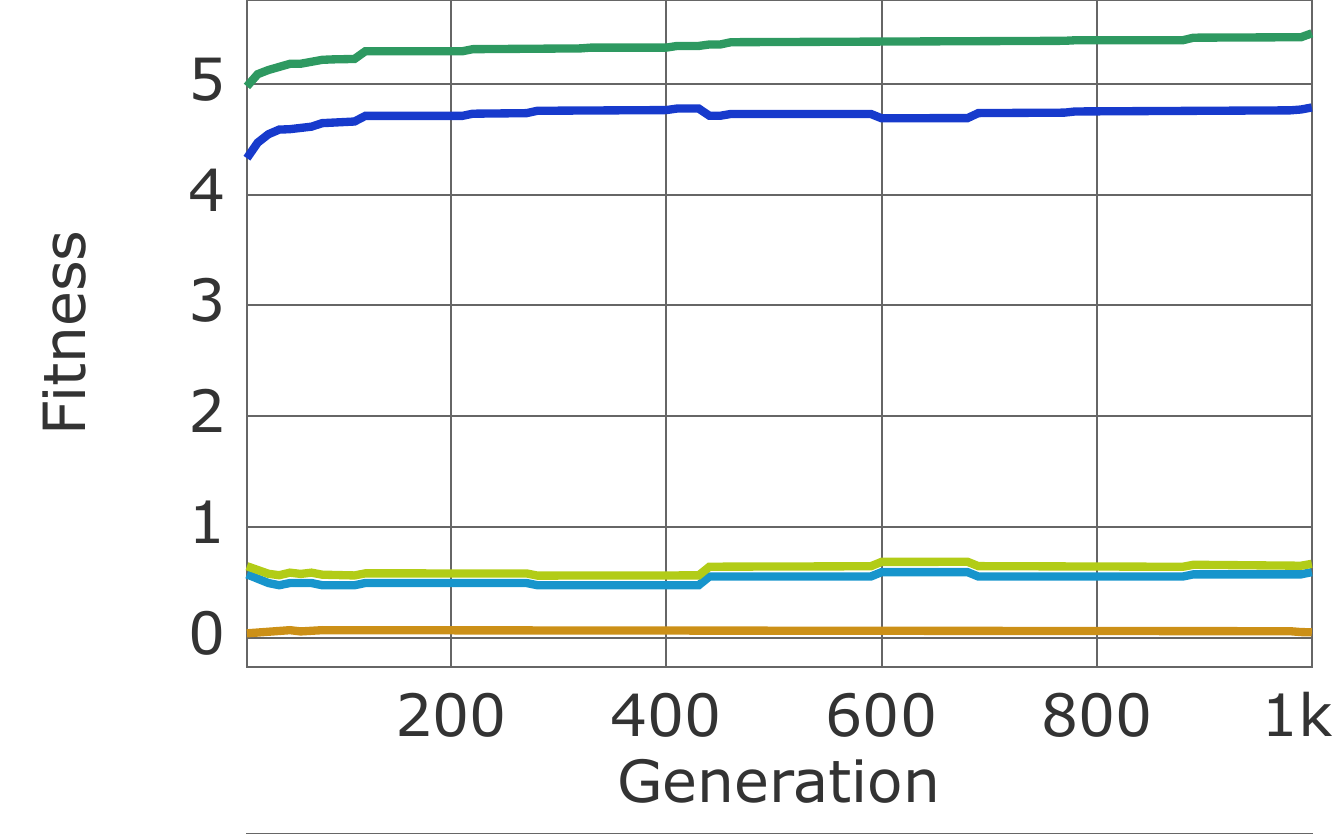}}\\ \vspace{0.4cm}
\subfloat[Population Analysis\label{fig:FetchReach-Population-SAC3M}]{\includegraphics[width=0.5\linewidth]{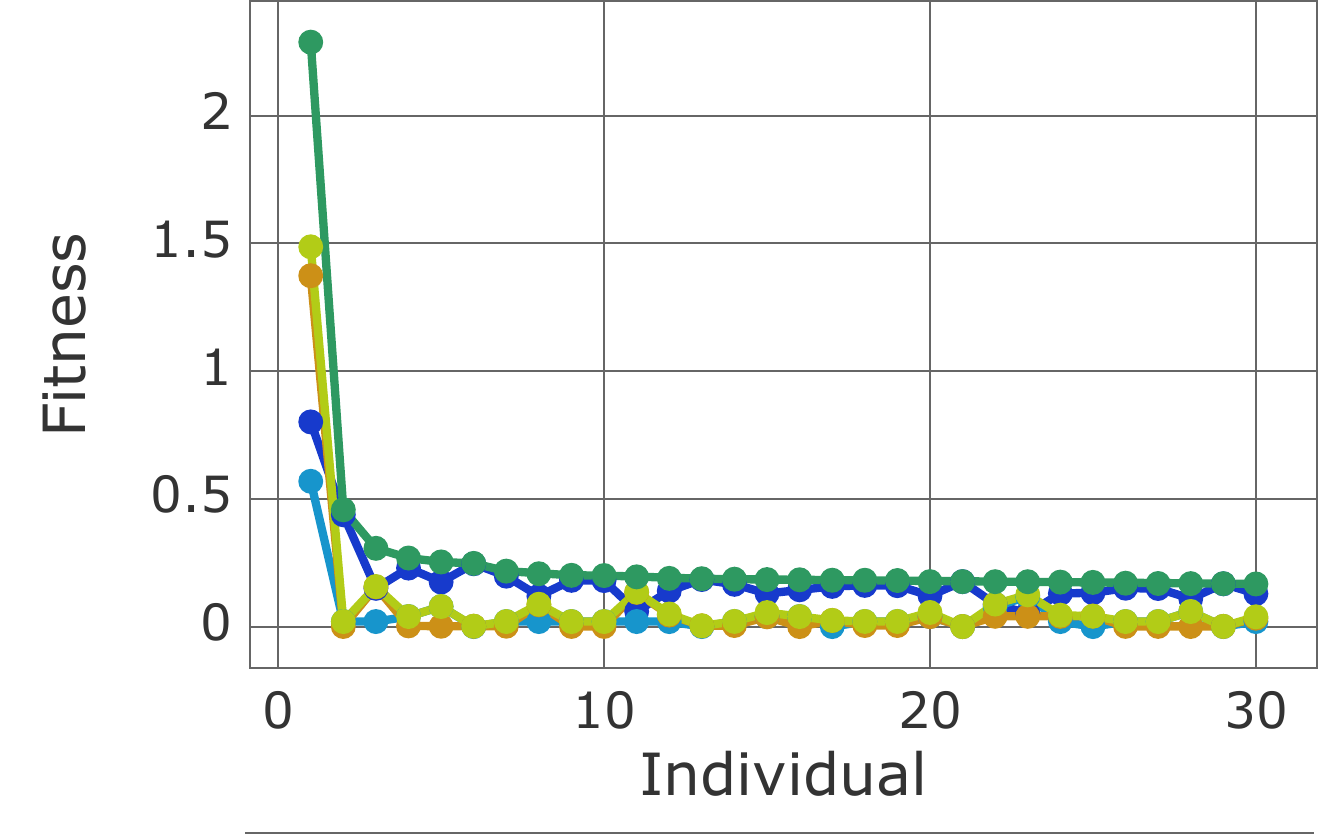}}
\subfloat[Generation Analysis\label{fig:FetchReach-Generations-SAC3M}]{\includegraphics[width=0.5\linewidth]{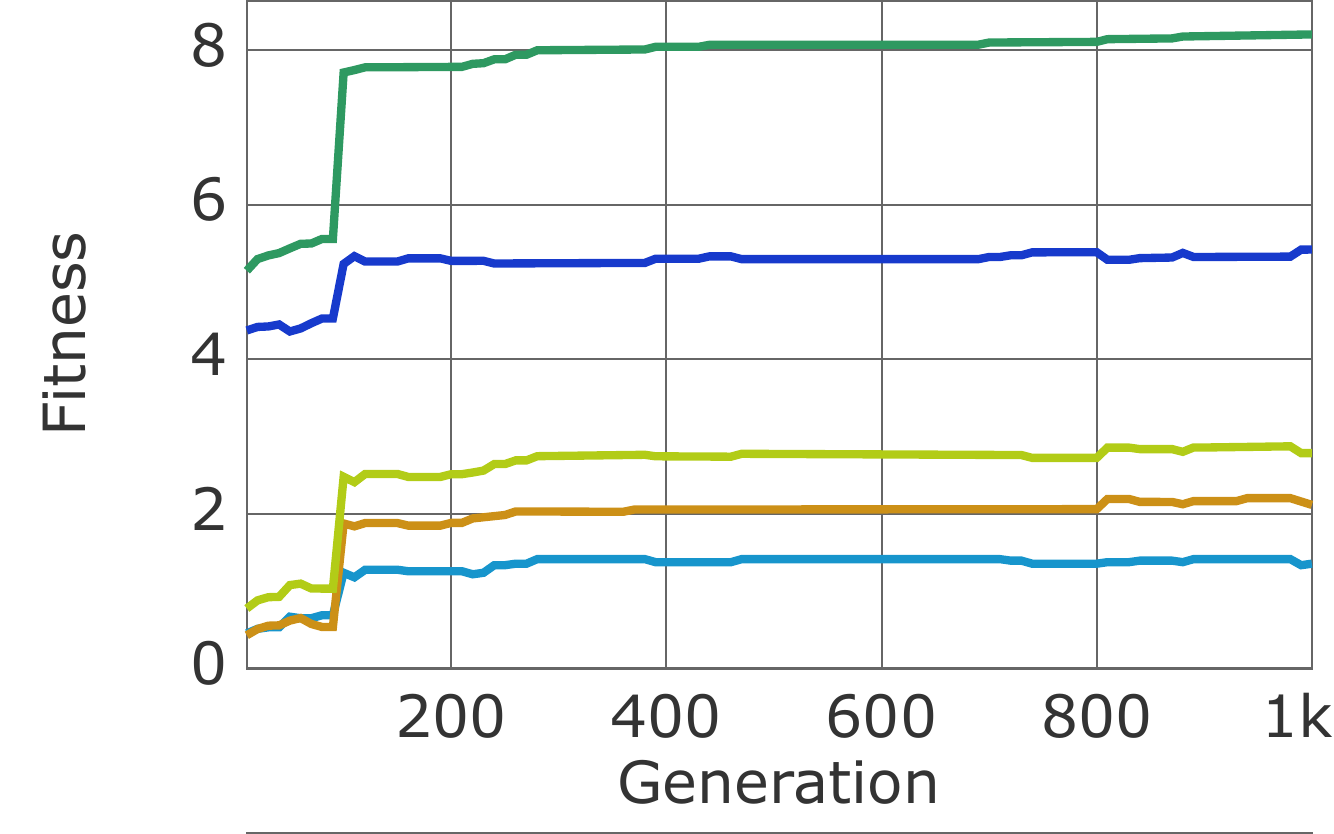}}\\ \vspace{0.4cm}
\subfloat[Population Analysis\label{fig:FetchReach-Population-SAC5M}]{\includegraphics[width=0.5\linewidth]{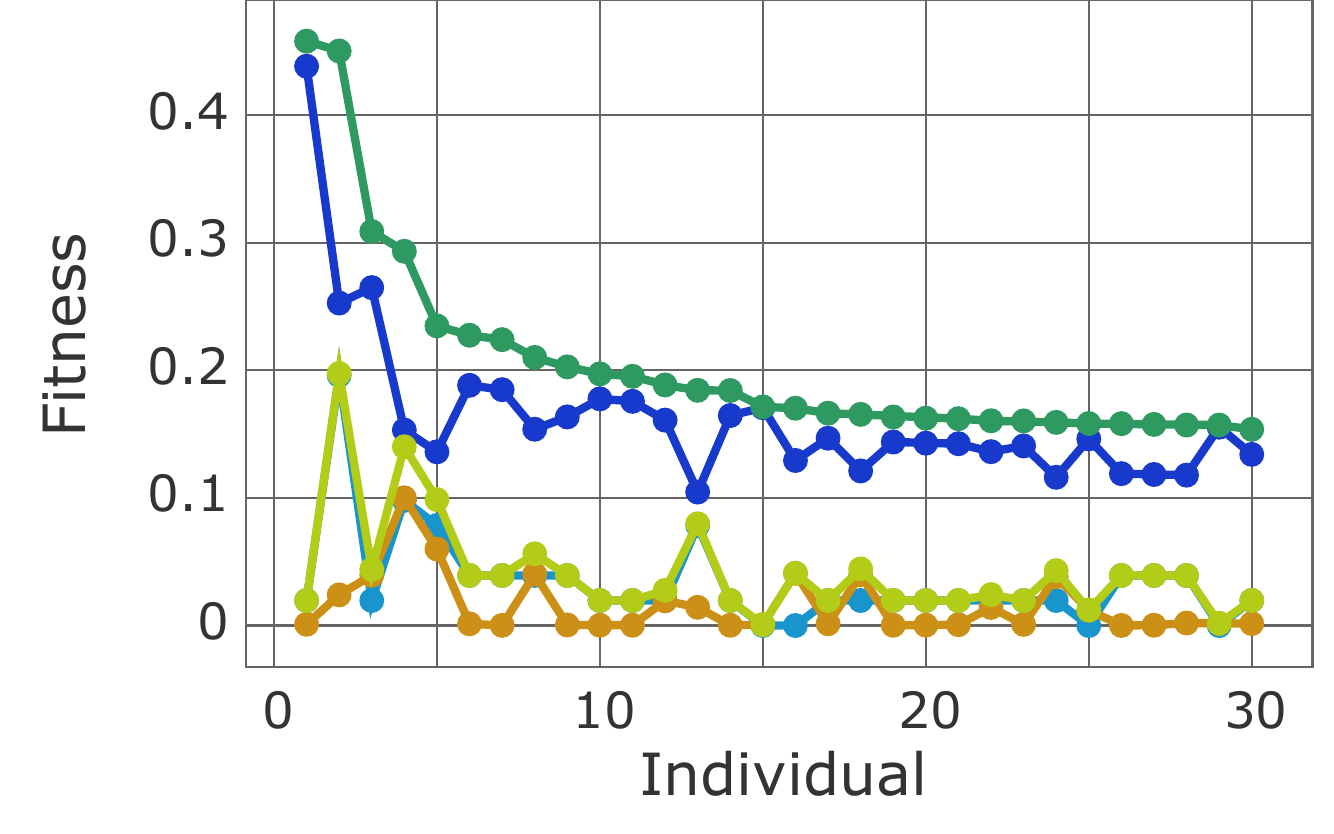}}
\subfloat[Generation Analysis\label{fig:FetchReach-Generations-SAC5M}]{\includegraphics[width=0.5\linewidth]{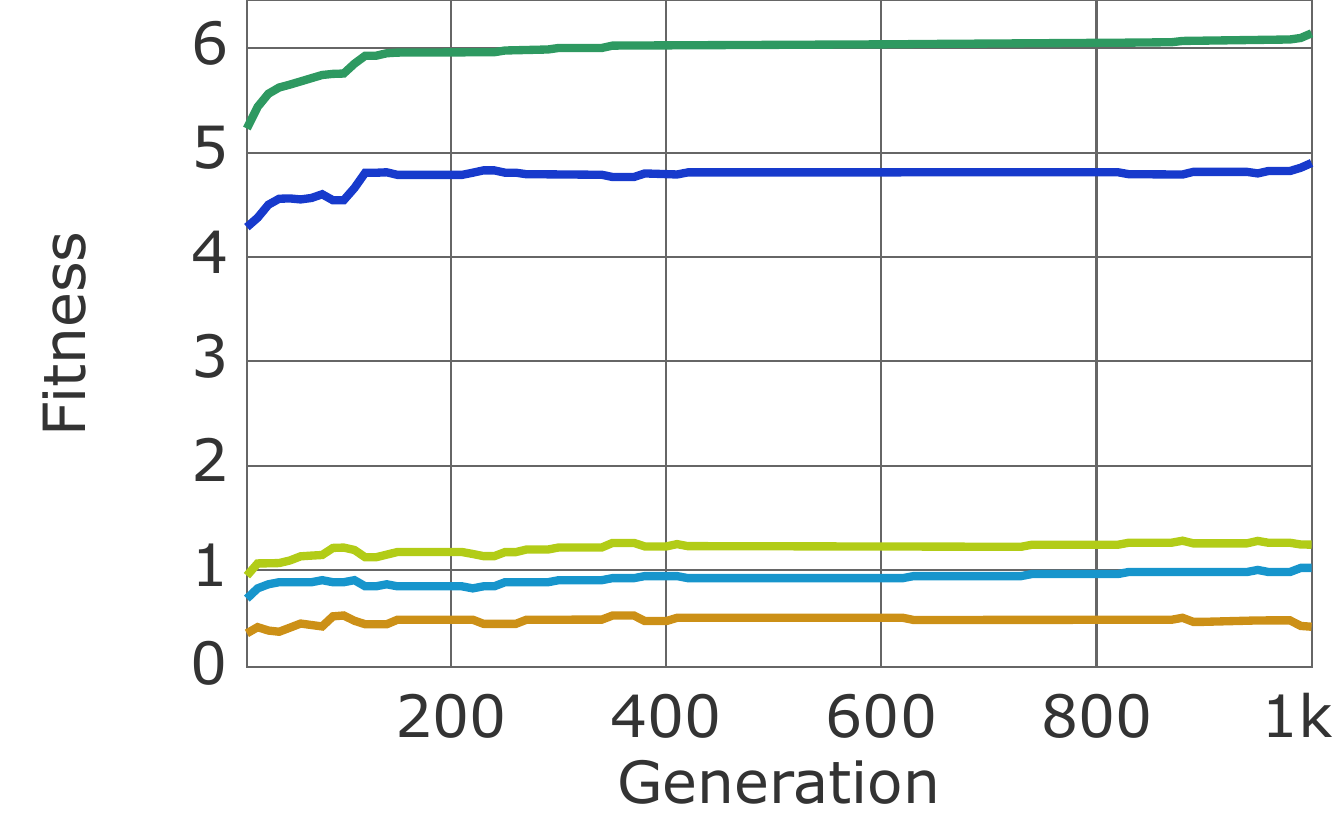}}
\vspace{-0.1cm}\caption{FetchReach \textit{JointFitness} Analysis\label{fig:FetchReach-Fitness}}
\Description{FetchReach \textit{JointFitness} Analysis}
\end{figure}

\end{document}